\documentclass[letterpaper, 10 pt, conference]{ieeeconf}  % Comment this line out if you need a4paper

\IEEEoverridecommandlockouts                              % This command is only needed if 
\usepackage{graphics} % for pdf, bitmapped graphics files
\usepackage{epsfig} % for postscript graphics files
\usepackage{times} % assumes new font selection scheme installed
\usepackage{amsmath} % assumes amsmath package installed
\usepackage{amssymb}  % assumes amsmath package installed
\usepackage{multicol}
\usepackage{multirow}
\usepackage{booktabs}
\usepackage{pifont}         % xmark, cmark
\usepackage{colortbl}
\usepackage[table]{xcolor} % for colors
\usepackage[export]{adjustbox} % valign=c
\usepackage{hyperref}

\newcommand \inline {\noindent\textbf}
\def \ours {TUN3D}
\newcommand{\cmark}{\ding{51}}
\newcommand{\xmark}{\ding{55}}

\title{\LARGE \bf
TUN3D: Towards Real-World Scene Understanding from \\ Unposed Images
}

\author{\textbf{Anton Konushin}\textsuperscript{1}, \textbf{Nikita Drozdov}\textsuperscript{1}, \textbf{Bulat Gabdullin}\textsuperscript{2}, \textbf{Alexey Zakharov}\textsuperscript{1}, \textbf{Anna Vorontsova},\\ \textbf{Danila Rukhovich}\textsuperscript{3\dag}, \textbf{Maksim Kolodiazhnyi}\textsuperscript{1}\\ \textsuperscript{1}Lomonosov Moscow State University; \textsuperscript{2}Higher School of Economics; \textsuperscript{3}Institute of Mechanics, Armenia}

\begin{document}

\maketitle
\thispagestyle{empty}
\pagestyle{empty}

%%%%%%%%%%%%%%%%%%%%%%%%%%%%%%%%%%%%%%%%%%%%%%%%%%%%%%%%%%%%%%%%%%%%%%%%%%%%%%%%
\begin{abstract}

\let\thefootnote\relax\footnotetext{\textsuperscript{\dag}Corresponding author: rukhovich@mechins.sci.am}
Layout estimation and 3D object detection are two fundamental tasks in indoor scene understanding. When combined, they enable the creation of a compact yet semantically rich spatial representation of a scene. Existing approaches typically rely on point cloud input, which poses a major limitation since most consumer cameras lack depth sensors and visual-only data remains far more common. We address this issue with \ours{}, the first method that tackles joint layout estimation and 3D object detection in real scans, given multi-view images as input, and does not require ground-truth camera poses or depth supervision. Our approach builds on a lightweight sparse-convolutional backbone and employs two dedicated heads: one for 3D object detection and one for layout estimation, leveraging a novel and effective parametric wall representation. Extensive experiments show that \ours{} achieves state-of-the-art performance across three challenging scene understanding benchmarks: (i) using ground-truth point clouds, (ii) using posed images, and (iii) using unposed images. While performing on par with specialized 3D object detection methods, \ours{} significantly advances layout estimation, setting a new benchmark in holistic indoor scene understanding. Code is available at \url{https://github.com/col14m/tun3d}.

\end{abstract}

%%%%%%%%%%%%%%%%%%%%%%%%%%%%%%%%%%%%%%%%%%%%%%%%%%%%%%%%%%%%%%%%%%%%%%%%%%%%%%%%
\section{INTRODUCTION}

Indoor scene understanding is a long-standing computer vision task, with applications in robotics, AR/VR, interior design, real estate and property inspection. Many real-world scenarios require only a compact yet informative structural description of the scene: the room layout (walls, floor, ceiling) and the locations and categories of major objects. Being significantly lighter than dense 3D reconstruction, this representation is especially beneficial for applications running on a device.

Recent approaches address 3D object detection~\cite{liu2021groupfree3d, rukhovich2022fcaf3d, rukhovich2023tr3d, xu2023nerfdet, huang2024nerfdets, zhu2024spgroup3d, huang2025nerfdetplusplus, kolodiazhnyi2025unidet3d}, layout estimation~\cite{yue2023roomformer, gao2023omnipq}, or even explore joint prediction of layout and objects from point clouds~\cite{chen2022pq, avetisyan2024scenescript, mao2025spatiallm}. Arguably, a joint model could benefit from shared reasoning about the structure and semantics of a scene, but existing methods are either 
extremely slow~\cite{avetisyan2024scenescript, mao2025spatiallm}, or still far behind single-task methods in accuracy~\cite{chen2022pq}. On the contrary, we build our model on top of a real-time 3D object detection model, achieving state-of-the-art performance with impressive latency.

Input data is another critical issue in the 3D scene understanding. Using point clouds at inference time requires either depth sensors and/or accurate multi-view reconstruction. This limits the applicability of point cloud-based methods in scenarios where only video is available, e.g., on customer devices that are equipped with neither depth sensors nor trackers, or for processing prerecorded videos. In this paper, we investigate input data modalities in the 3D scene reconstruction context and move beyond point clouds towards images with camera poses and even unposed images. As a result of this study, we present \ours{}, the first method of layout estimation and 3D object detection from real-world images without access to depth or camera poses. 

Our contribution is as follows: 
\begin{itemize}
    \item We show that data requirements in 3D scene understanding can be relaxed from point clouds to multi-view images -- with and even without camera poses;
    \item We propose a compact yet expressive layout parameterization and a new architecture for joint layout estimation and 3D object detection;
    \item We establish a new state-of-the-art in scene understanding across three challenging scenarios: from ground truth point clouds, posed multi-view images, and unposed multi-view images.
\end{itemize}

\section{RELATED WORK}

\subsection{Scene Understanding from Point Clouds}

\begin{figure*}[t!]
    \begin{center}
        \includegraphics[width=\linewidth]{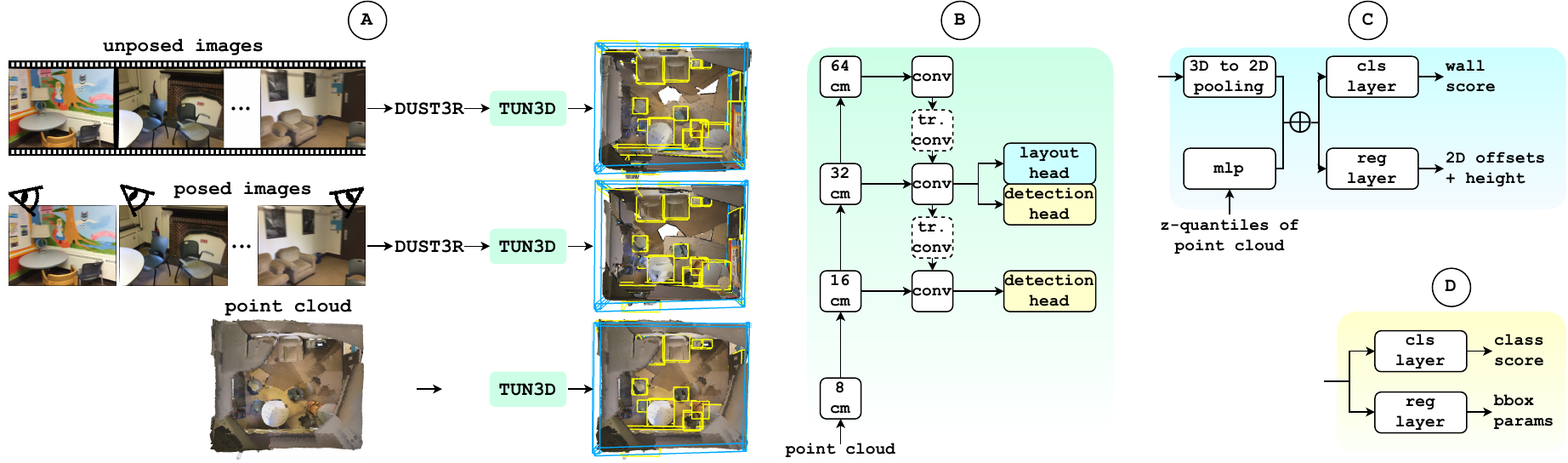}
    \end{center}
    \caption{(A) \ours{} can flexibly process various inputs: unposed images, posed images, and point clouds. (B) \ours{} model is constructed of a 3D sparse-convolutional backbone and neck, followed by two task-specific heads. (C) The novel layout head predicts wall scores and regresses wall parameters for each wall comprising the layout. (D) The detection head outputs object class scores and coordinates of a 3D bounding box of an object.}
\label{fig:scheme}
\end{figure*}

Point cloud-based 3D object detection methods can be categorized into voting-based, transformer-based, and sparse convolutional. Voting-based methods, from the seminal VoteNet~\cite{qi2019votenet} to recent SPGroup3D~\cite{zhu2024spgroup3d}, follow a bottom-up paradigm, grouping processed points into object candidates. Transformer-based methods, including the first-in-class GroupFree3D~\cite{liu2021groupfree3d} and the most recent UniDet3D~\cite{kolodiazhnyi2025unidet3d}, use a transformer encoder to predict a set of objects. Methods based on sparse 3D convolutions, such as GSDN~\cite{gwak2020gsdn}, FCAF3D~\cite{rukhovich2022fcaf3d} and its follow-up TR3D~\cite{rukhovich2023tr3d}, balance speed, accuracy, and scale well to larger scenes, so we develop our \ours{} following this paradigm.

Layout estimation is another long-lasting scene understanding task that has been addressed solely by such methods as Omni-PQ~\cite{gao2023omnipq}, RoomFormer~\cite{yue2023roomformer}, or coupled with 3D object detection in SceneCAD~\cite{avetisyan2020scenecad} and PQ-Transformer~\cite{chen2022pq}.

Recent emergence of large language models boosted the scene understanding research. SceneScript~\cite{avetisyan2024scenescript} trains a transformer model to generate procedural scene descriptions in a language created specifically for this task, while SpatialLM~\cite{mao2025spatiallm} outputs Python code for scene generation. Both SceneScript and SpatialLM are trained with synthetic data solely, and no layout estimation results are shown on real scans.

\subsection{Scene Understanding from Posed Images}

Numerous methods can process visual information rather than explicit scene geometry but gain such an ability by utilizing depth for training. ImGeoNet~\cite{tu2023imgeonet} learns geometry from multi-view images with depth supervision. 3DGeoDet~\cite{zhang20253dgeodet} uses depth to estimate voxel occupancy, while NeRF-based methods, including NeRF-DetS~\cite{huang2024nerfdets}, NeRF-Det++~\cite{huang2025nerfdetplusplus} and GO-N3RDet~\cite{li2025gon3rdet}, implement various ways of multi-view feature fusion with depth guidance.

Depth-free 3D object detection from posed images is an emerging research topic, launched with the publication of ImVoxelNet~\cite{rukhovich2022imvoxelnet} that first tackled 3D object detection from multi-view inputs in an end-to-end manner. Later, NeRF-Det~\cite{xu2023nerfdet} exploite NeRF's ability to infer 3D geometry from sole visual inputs for depth-free 3D object detection. MVSDet~\cite{xu2024mvsdet} utilizes plane sweep for geometry-aware 3D object detection, applying probabilistic sampling and a soft weighting mechanism for feature lifting in the absence of explicit depth. SceneScript~\cite{avetisyan2024scenescript} also provides a version capable of handling posed images, but neither layout estimation nor 3D object detection is reported on real-world images.

\subsection{Scene Understanding from Unposed Images}

Single-view layout estimation~\cite{nie2020total3dunderstanding} and 3D object detection~\cite{rukhovich2022imvoxelnet} approaches are flexible but limited in scene coverage, making the predictions hardly usable in real-world applications. Panoramic images provide a more complete view of the scene and are widely used for layout estimation~\cite{wang2021led2, jiang2022lgt, dong2024panocontext}. However, they are inherently limited to a single viewpoint, which restricts object coverage and often leads to occlusion issues. 

Other scene understanding tasks, such as 3D visual grounding, 3D dense captioning, and 3D question answering, are now being solved with LLMs with visual capabilities that take images as inputs~(\cite{chen2024spatialvlm, zhu2024llava, cheng2024spatialrgpt, zheng2025video, zhi2025lscenellm}). Point cloud-based SpatialLM~\cite{mao2025spatiallm} is combined with Mast3R~\cite{leroy2024mast3r} for video processing, and this combination is proved to successfully handle synthetic data. To the best of our knowledge, there are no prior methods of either layout estimation or 3D object detection taking multi-view unposed images as inputs and working with real-world data. In this paper, we aim to close this gap with \ours{}.

\section{SCENE UNDERSTANDING FROM POINT CLOUD}

\subsection{Problem Formulation}

In \ours{}, we formulate scene understanding as proposed in PQ-Transformer~\cite{chen2022pq}, and predict layouts and detect 3D objects jointly with a single model, given a colored point cloud as an input. More formally, \ours{} model awaits a point cloud $P=\{p_i\}_{i=1}^N \subset \mathbb{R}^6$, where each point $p_i=(x_i,y_i,z_i,r_i,g_i,b_i)$ is described with its coordinates in the 3D space and RGB color.

Detected 3D objects are parameterized as $\mathcal{O}=\{(b_k,c_k)\}_{k=1}^{K}$, where $c_k\in\{1,\dots,C\}$ denotes object categories and $b_k$ stands for the spatial parameters of a 3D bounding box. $b_k=(t_k,s_k)$, where $t_k\in\mathbb{R}^3$ is the center of a 3D bounding box and $s_k\in\mathbb{R}^3_{+}$ are sizes along the $x,y,z$-axes.

Layout is defined as a set of walls $\mathcal{W}=\{w_\ell\}_{\ell=1}^{L}$, where each wall $w_\ell = (q_{\ell,1},q_{\ell,2},q_{\ell,3},q_{\ell,4})$ is specified by 3D coordinates of its four corners  $q_{\ell,j}\in\mathbb{R}^3$, ordered clockwise.

\subsection{Network Architecture}

Our fully-differentiable model is built of a backbone, neck, and two heads addressing 3D object detection and layout estimation, respectively (Fig.~\ref{fig:scheme}). Below, we discuss these components architecture-wise and describe techniques used during the training and inference phases.

\inline{Backbone} is a 3D sparse high-dimensional version of ResNet, first introduced in GSDN~\cite{gwak2020gsdn} and used in FCAF3D~\cite{rukhovich2022fcaf3d}. We use an optimized version described in TR3D~\cite{rukhovich2023tr3d}. First, an input point cloud is voxelized with a voxel size of 2 cm. Then, four residual blocks of sparse 3D convolutional layers transform the voxel space into 8 cm, 16 cm, 32 cm, and 64 cm-sized spatial grids. The maximum number of channels in all sparse convolutional layers is upper-limited to 128 for the sake of efficiency, since larger values compromise inference speed. 

\begin{figure*}[t!]
    \centering \small
    \includegraphics[width=\linewidth]{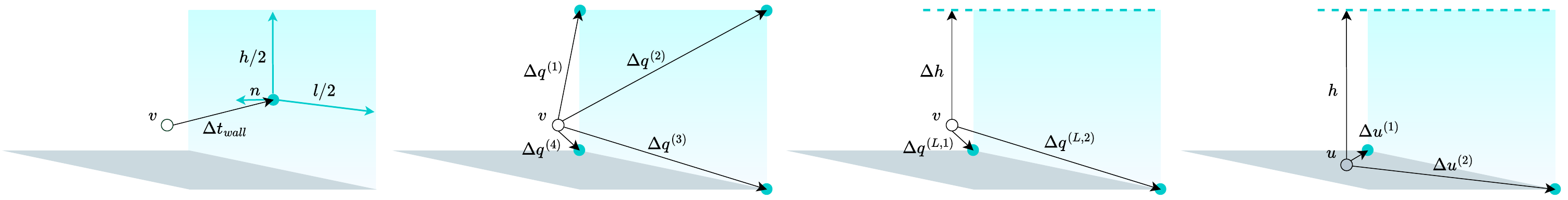}
    \begin{minipage}[t]{0.25\linewidth}
        \centering
        PQ~\cite{chen2022pq}
    \end{minipage}%
    \begin{minipage}[t]{0.25\linewidth}
        \centering
        4$\times$3D offsets
    \end{minipage}%
    \begin{minipage}[t]{0.25\linewidth}
        \centering
        2$\times$3D offsets + height
    \end{minipage}%
    \begin{minipage}[t]{0.25\linewidth}
        \centering
        \textbf{2$\times$2D offsets + height, ours}
    \end{minipage}
    \caption{Different wall parameterizations.}
    \label{fig:wall_param}
\end{figure*}

\inline{Neck} aggregates 3D voxel features from four residual levels of the backbone. Features at each level are processed with one sparse generative transposed 3D convolution and one sparse 3D convolution. The well-known issue of standard convolutions is that they might downscale the visibility field too aggressively. To prevent loss of spatial information, we apply generative convolutional layers: they increase the number of voxels, hence expanding the covered area, so that object candidates beyond the current visibility field can still be processed. We use generative convolutions only at levels of 64 cm and 32 cm. 

\inline{Detection head} in \ours{} is similar to the one used in TR3D~\cite{rukhovich2023tr3d}. Specifically, it consists of two linear layers stacked sequentially. The weights are shared across 32 cm and 16 cm feature levels. The head returns a set of 3D locations $\hat{\mathcal V}=\{\hat v_j\}_{j=1}^{J}$, and for each location $\hat v_j\in\hat{\mathcal V}$, it predicts class logits $\tilde z_j\in\mathbb{R}^{C}$, offset of the center of an object $\Delta t_j\in\mathbb{R}^{3}$, and log-sizes of its 3D bounding box $\tilde s_j\in\mathbb{R}^{3}$. The canonical representation of the predicted 3D bounding box is derived as $t_j = \hat v_j + \Delta t_j$, $s_j = \exp(\tilde s_j) \in \mathbb{R}^3_{+}$,
while the resulting class probabilities are calculated as $p_{jc}=\sigma(\tilde z_{jc})$.

\inline{Layout head} takes features from $\hat{\mathcal V}=\{\hat v_j\}_{j=1}^{J}$ at the 32-cm level as inputs and produces a layout as a set of walls. For each wall, wall logit $\alpha_j\in\mathbb{R}$ is predicted with the classification layer, so that the final wall probability score is $p^\text{wall}_j=\sigma(\alpha_j)$. Wall parameters are estimated with a novel regression layer, discussed below.

\subsection{Wall Parameterization}

Since we tackle layout estimation in the formulation of PQ-Transformer~\cite{chen2022pq}, we use the same wall parameterization. However, in our experiments it appeared to be suboptimal, while other parameterizations delivered better quality. Below, we elaborate on possible alternatives, review existing approaches, and formalize novel ways to define a single wall.

\inline{PQ parameterization}~\cite{chen2022pq} defines a wall with eight parameters, namely, an offset from the location $v_j$ to the center of the wall $\Delta t^{\text{wall}}_j\in\mathbb{R}^3$, wall length $\ell_j\in\mathbb{R}_{+}$, wall height $h_j\in\mathbb{R}_{+}$, and a normal to the wall plane $n_j\in\mathbb{R}^3$, $\|n_j\|_2=1$. Accordingly, the wall center in world coordinates is calculated as $c_j = \hat v_j + \Delta t^{\text{wall}}_j$, and the four corners $q_{j,1:4}$ are derived from $c_j,\ell_j,h_j,n_j$ trivially.

\inline{4$\times$3D offsets} 
Arguably the most straightforward way to define a wall is to simply specify its four corners. This non-parametric wall representation would contain 3D offsets to all corners, $\Delta q^{(k)}_j\in\mathbb{R}^3$, $k=1,\dots,4$, totaling 12 values. Respectively, the wall corners in world coordinates can be restored by summing the location position with offsets, i.e., $q^{(k)}_j=\hat v_j + \Delta q^{(k)}_j$.

\inline{2$\times$3D offsets + height}.
Assuming that walls have constant height, we can reduce the number of parameters to just seven: in such a case, we can specify only two 3D offsets to the lower (floor-touching) corners of the wall $\Delta q^{(L,1)}_j,\Delta q^{(L,2)}_j\in\mathbb{R}^3$, and set a relative height $\Delta h_j\in\mathbb{R}_{+}$. 
Lower wall corners can then be derived as $q^{(L,m)}_j=\hat v_j + \Delta q^{(L,m)}_j$ for $m\in\{1,2\}$. Upper wall corners are calculated as $q^{(U,m)}_j = q^{(L,m)}_j + \Delta h_j\, e_z$, given that $e_z$ is a global up axis.

\inline{2$\times$2D offsets + height (ours)} In all three wall parameterizations above, predicted 3D offsets are not mutually constrained, which might result in a malformed scene geometry. But what if we enforce more rigidity by reducing the degrees of freedom even further -- will it help to obtain cleaner geometry and make the model more robust and general? Taking insights from outdoor 3D object detection approaches~\cite{lang2019pointpillars, rukhovich2022imvoxelnet}, we perform dimensional reduction and define the layout estimation task in the bird's-eye-view (BEV) plane instead of 3D space. The intuition behind this trick is the same as in the outdoor scenario: the cars cannot be stacked on top of each other -- so neither can the walls. 

The proposed dimensionality reduction is incorporated into the architecture of the layout head (Fig.~\ref{fig:scheme}, C). Particularly, 3D features yielded by the neck are first projected onto the floor plane via average pooling. As a result of the pooling procedure, important information about the height of walls is being lost, so later on we enrich those floor-projected features with missing spatial information. To this end, we calculate heights of all points in a scene, which are naturally their $z$ coordinates, since the floor is zero-aligned, and estimate $z$-quantiles. $z$-quantiles are then encoded with a small MLP into a single vector encapsulating height distribution in the scene. This scene-level vector is concatenated to all projected features at predicted locations $\hat u_j:=(\hat x_j,\hat y_j)$. 

Eventually, the wall is encoded with five parameters: 2D offsets to the two lower wall corners
$\Delta u^{(1)}_j,\Delta u^{(2)}_j\in\mathbb{R}^2$ and height $h_j\in\mathbb{R}_{+}$.
The lower wall corners could be computed as $q^{(L,m)}_j=\big(\hat u_j+\Delta u^{(m)}_j,0\big)$, and upper wall corners are $q^{(U,m)}_j=q^{(L,m)}_j+h_j\,e_z$, $m\in\{1,2\}$.

\subsection{Training}

To calculate loss during the training, predicted objects should be assigned to ground truth objects. The matching rules do not need to be the same for both tasks being solved; actually, we apply different matching strategies for objects and walls.

\inline{Location-object assignment} is performed to couple 3D locations $\{\hat v_j\}$ with ground truth objects $\mathcal{O}$. Following the matching strategy in~\cite{rukhovich2023tr3d}, we pre-define the head level for each object category: typically large objects (e.g., bed or sofa) are processed at the third level (32 cm), and smaller ones (e.g., chair or nightstand) are handled at the second (16 cm). Within a feature level specified, each ground truth object is assigned six locations nearest to its center.

\inline{Location-wall assignment} aims to establish correspondence between 3D locations $\{\hat v_j\}$ (or 2D locations $\{\hat u_j\}$) and ground truth walls $\mathcal{W}$. For all wall parameterizations, we adopt the assignment strategy for "large" objects: assigning a wall to the six nearest 3D locations (or their 2D projections onto the floor plane) predicted at the 32-cm feature level. A similar mechanism is integrated for wall parameterizations based on both 3D offsets and 2D floor projections.

\inline{Loss} is multi-component, where each component contributes to training a specific output layer in heads. Namely, classification in the 3D object detection head is guided with a focal loss, regression of 3D bounding box parameters is being trained with DIoU loss. To penalize erroneous layouts, the focal loss is applied to the outputs of the wall classification layer, and L1 loss is estimated for wall parameters. The overall training loss is thus calculated as follows:
\[
\mathcal{L} = \mathcal{L}_{\text{focal}}^{\text{det}} + \mathcal{L}_{\text{DIoU}}^{\text{det}} + \mathcal{L}_{\text{focal}}^{\text{layout}} + \mathcal{L}_{\text{L1}}^{\text{layout}}
\]

\section{TOWARDS SCENE UNDERSTANDING FROM UNPOSED IMAGES}

In this section, we explore using fewer ground truth modalities for scene understanding during both training and inference. Relaxing the requirements for input data may open new possibilities for running scene understanding applications on customer devices that are not equipped with depth sensors or trackers -- or process pure visual data collected with casual cameras. Respectively, besides point cloud being the most informative and resource-intensive modality, we experiment images with camera poses, and even unposed images as inputs.

\subsection{Posed Images}

In the pose-aware scenario, our method accepts a set of images $\{I_m\}_{m=1}^M$, $I_m\in\mathbb{R}^{H\times W\times 3}$ along with camera intrinsics $K_m\in\mathbb{R}^{3\times 3}$ and camera extrinsics $T_m\in SE(3)$. In real applications, camera poses may be sourced using either specialized hardware (inertial measurement units) or integrated software (visual trackers).

As discussed above, we are already able to process point clouds. The missing part of the puzzle is hence converting images with poses into a point cloud. To this end, we employ dense structure-from-motion methods, namely DUSt3R~\cite{wang2024dust3r} which has certain advantages over similar approaches. First, it can operate not only in pose-agnostic, but also in pose-aware mode, accepting ground truth poses as auxiliary inputs. Respectively, it is applicable in both posed-image and unposed-image scenarios, making the entire framework flexible and laconic at the same time.
Apart from that, using DUSt3R also allows preserving methodological purity of our experimental protocol and avoiding data leakage, since it was not trained on ScanNet, contrary to some competing methods (including its famous follow-up, Mast3R~\cite{leroy2024mast3r}).

Being a dense SfM method, DUSt3R estimates dense depth maps for given frames. Then, the original images and those estimated depths are fused using ground truth camera poses into a TSDF volume. Finally, a point cloud is extracted, -- the rest steps of the solution match the steps taken in the point cloud-based scenario.

\subsection{Unposed Images}

In the third scenario, our model is challenged to process a sole image collection $\{I_m\}_{m=1}^M$ without known camera intrinsics $K_m$ or extrinsics $T_m$. This formulation is relevant for data captured with most casual customer devices (e.g., smartphones or non-professional cameras), or prerecorded videos with missing capturing information.

DUSt3R reveals its full potential in this most challenging scenario, jointly predicting depth maps and camera parameters. Same as in the pose-aware case, those depth maps are then used for TSDF integration, but here we rely on estimated camera poses instead of ground truth ones.

%In both image-based setups, we do not use ground truth point clouds and reconstruct them instead. Still, ground truth annotations are needed for training and evaluation, since scans must be transformed into common coordinate space before computing the metrics. To estimate an affine transformation that aligns ground truth and predicted point clouds, we use ground truth and predicted camera poses of two first (time-wise) frames. Specifically, the first predicted pose is aligned with the first ground truth pose, giving the rotation and translation. The relative scale coefficient is derived as a ratio of distances between two camera poses in predicted and ground truth scans.

\section{EXPERIMENTS}

\begin{table*}[t!]
\centering
\caption{Results of layout estimation and object detection from various input modalities on ScanNet and S3DIS.}
\label{tab:results}
\begin{tabular}{llccccccccc}
\toprule
& \multirow{3}{*}{Method} & \multirow{3}{*}{Venue} & \multicolumn{2}{c}{Depth} & \multicolumn{3}{c}{ScanNet}  & \multicolumn{3}{c}{S3DIS} \\
\cmidrule(r{0.2em}){6-8} \cmidrule(l{0.2em}){9-11}
& &  & \multirow{2}{*}{Train} & \multirow{2}{*}{Test} & Layout & \multicolumn{2}{c}{Detection} & Layout & \multicolumn{2}{c}{Detection} \\
& & & & & F1 & mAP@0.25 & mAP@0.5 & F1 & mAP@0.25 & mAP@0.5 \\
\midrule
\multicolumn{10}{l}{\textit{GT point clouds}} \\
& GSDN~\cite{gwak2020gsdn}  & ECCV'20 & \cmark & \cmark & - & 62.8 & 34.8 & - &  47.8 & 25.1 \\
& TR3D~\cite{rukhovich2023tr3d}  & ICIP'22 & \cmark & \cmark & - & 72.0 & 58.1 & - & 72.5  & 57.2  \\
& SPGroup3D~\cite{zhu2024spgroup3d} & AAAI'24 & \cmark & \cmark & - & 74.3 & 59.6 & - & 69.2  & 47.2  \\
& UniDet3D~\cite{kolodiazhnyi2025unidet3d}  & AAAI'25 & \cmark & \cmark & - & \textbf{77.0} & \textbf{65.0} & - & 73.2  & 57.4  \\
& SceneCAD~\cite{avetisyan2020scenecad}  & ECCV'20 & \cmark & \cmark & 37.9 & - & - & - & -  & -  \\
& Omni-PQ~\cite{gao2023omnipq}  & ICRA'23 & \cmark & \cmark & 60.8 & - & - & - & -  & -  \\
& PQ~\cite{chen2022pq}  & ICRA'22 & \cmark & \cmark & 54.4 & 60.9 & 39.9 & 29.6 & 61.1  & 38.0  \\
& \textbf{\ours{}}  & - & \cmark & \cmark & \textbf{66.6} & 72.7 & 60.2 & \textbf{53.2} & \textbf{74.4} & \textbf{58.6}  \\
\midrule
\multicolumn{10}{l}{\textit{Posed images}} \\
& NeRF-Det++~\cite{huang2025nerfdetplusplus} & TIP'25 & \cmark & \xmark  & - & 53.3 & 30.0 & -  & -  & -  \\
& ImGeoNet~\cite{tu2023imgeonet}  & ICCV'23 & \cmark & \xmark  & -  & 54.8  & 28.4  & -   & -  & -  \\
& NeRF-DetS~\cite{huang2024nerfdets} & arXiv'24  & \cmark  & \xmark  & -  & 57.6  & 35.6 & -   & -  & -  \\
& GO-N3RDet~\cite{li2025gon3rdet}  & CVPR'25 & \cmark & \xmark  & - & 58.6 & 33.7 & - & - & -  \\
& 3DGeoDet~\cite{zhang20253dgeodet} & TMM'25  & \cmark & \xmark & - & 59.6 & 34.3 & - & - & - \\
\arrayrulecolor{gray}\cmidrule{2-11}\arrayrulecolor{black}
& ImVoxelNet~\cite{rukhovich2022imvoxelnet}  & WACV'22  & \xmark  & \xmark & -  & 46.7  & 23.4  & - & - & - \\
& NeRF-Det~\cite{xu2023nerfdet}  & ICCV'23 & \xmark  & \xmark   & -  & 53.5  & 27.4  & -     & -  & -  \\
%& NeRF-Det++~\cite{huang2025nerfdetplusplus} & arXiv'24 & \xmark & \xmark  & - & 53.9 & 30.0 & -  & -  & -  \\
& MVSDet~\cite{xu2024mvsdet} & NeurIPS'24 & \xmark  & \xmark  & -  & 56.2 & 31.3 & - & - & - \\
& DUSt3R $\rightarrow$ PQ  & -  & \xmark & \xmark  & 44.1 & 50.3 & 27.7 &  10.2 & 27.0 & 5.0 \\
& DUSt3R $\rightarrow$ \textbf{\ours{}} & -  & \xmark  & \xmark & \textbf{55.2} & \textbf{57.4}  & \textbf{35.6}  &  \textbf{37.9} &  \textbf{34.8} & \textbf{13.4} \\
\midrule
\multicolumn{10}{l}{\textit{Unposed images}} \\
& DUSt3R $\rightarrow$ PQ & -  & \xmark & \xmark & 39.7 & 39.1 & 16.7 & 5.0 & 10.9 & 1.3 \\
& DUSt3R $\rightarrow$ \textbf{\ours{}} & - & \xmark & \xmark  & \textbf{46.5} & \textbf{44.0} & \textbf{20.7}  & \textbf{20.8} & \textbf{11.0} & \textbf{2.2} \\
\bottomrule
\end{tabular}
\end{table*}

\subsection{Datasets}

\inline{ScanNet}~\cite{dai2017scannet} is a widely used real-world dataset with 1201 scans in the training subset and 312 in the validation part. Following ~\cite{qi2019votenet}, we calculate axis-aligned 3D bounding boxes from semantic per-point labels. SceneCAD~\cite{avetisyan2020scenecad} further extends ScanNet with 3D layouts that we use in our experiments.

\inline{ARKitScenes}~\cite{baruch2021arkitscenes} is an RGB-D dataset containing 4493 training scans and 549 validation scans. The original dataset does not provide ground truth layouts. The validation subset was annotated in Omni-PQ~\cite{gao2023omnipq}, while the training part remains unlabeled; therefore, ARKitScenes is only used as a benchmark in cross-dataset experiments.  

\inline{S3DIS}~\cite{armeni2016s3dis} contains 272 scenes captured in six areas. Following the standard experimental protocol for object detection, we test on Area 5 and train on the rest areas, reporting detection accuracy for five semantic categories. We generate layout annotation by ourselves, calculating bounds of each wall instance. 

\inline{Structured3D}~\cite{zheng2020structured3d} is a large-scale synthetic dataset of 3.5K house designs created by professional designers along with ground truth 3D structure annotations and photo-realistic rendered images. The authors of ~\cite{yue2023roomformer} enriched Structured3D with structural elements, namely walls, windows, and doors, annotated in the floorplan. In our experiments, we use the SpatialLM~\cite{mao2025spatiallm} layout annotation created by lifting those floorplan annotations into the 3D space.

\subsection{Implementation Details}

\inline{Architecture.} The backbone and the neck of our model are the same as in TR3D~\cite{rukhovich2023tr3d}. From the architectural perspective, our main novelty is the layout head. In the case of using our novel wall parameterization, the information about spatial dimensions of a scene is distilled into a single vector. Precisely, we calculate 10 $z$-quantiles and encode them using a three-layer MLP with a ReLU into a vector of size 40. After that, this vector is concatenated to 128-channel 2D features passed from the neck so that the floor-projected representation contains 128 + 40 = 168 channels.

\inline{Training.} The training procedure follows the default mmdetection's learning schedule. We employ the Adam optimizer with an initial learning rate of 0.001 and weight decay of 0.0001. To control the size of input scenes, we sample a maximum of 100,000 points per scene. All experiments were conducted using a single Nvidia H100 GPU.

\inline{Inference.} During the inference, we generate excessive predictions and suppress the redundant ones using NMS, processing objects and walls separately. For objects, a positive match is found if the 3D IoU between the ground truth and predicted bounding box exceeds 0.5. Walls are matched if the maximum pairwise distance between all four ground truth and predicted corners of a wall falls under 75 cm.

\subsection{Metrics} 

To measure detection quality, we use mean average precision (mAP) under IoU thresholds of 0.25 and 0.5 as a metric.

According to the standard protocol, layout accuracy is assessed with an F1 score for ScanNet~\cite{dai2017scannet}, S3DIS~\cite{armeni2016s3dis}, and ARKitScenes~\cite{baruch2021arkitscenes} datasets, with walls being matched based on maximum corner-to-corner distance. The Structured3D~\cite{zheng2020structured3d} benchmark uses a variant of the F1 score, where matching is based on projection into the floorplan: specifically, two walls are matched if the IoU of their projections exceeds a given threshold; 0.25 and 0.5 are selected to mimic the evaluation protocol of object detection. 

\begin{figure*}[t!]
    \centering \small
    \begin{tabular}{cccccccc}
     & GT point clouds & Posed RGB & RGB-only & & GT point clouds & Posed RGB & RGB-only \\
     \rotatebox[origin=c]{90}{PQ~\cite{chen2022pq}} & 
     \includegraphics[width=0.12\linewidth, valign=c]{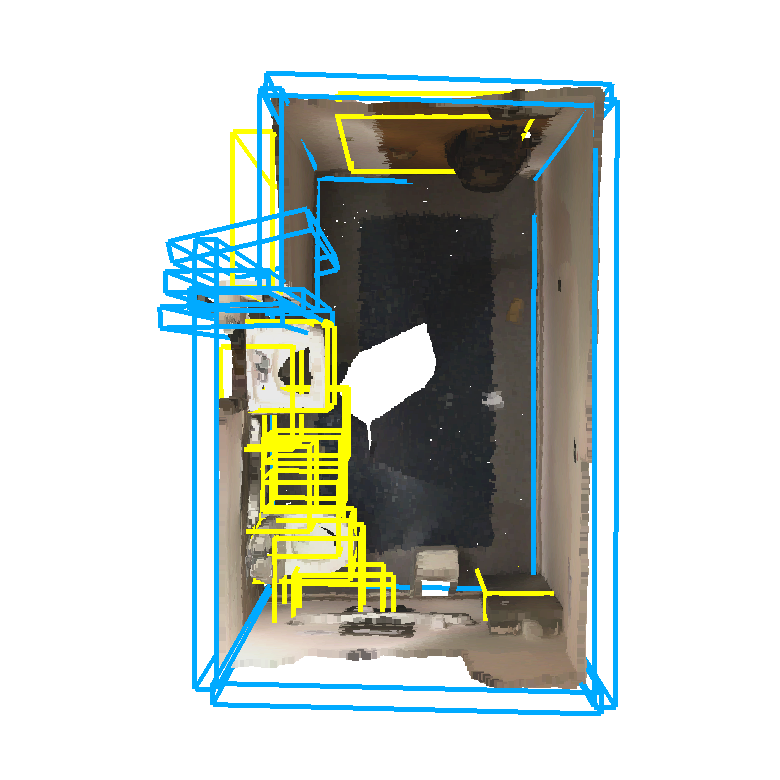} &
     \includegraphics[width=0.12\linewidth, valign=c]{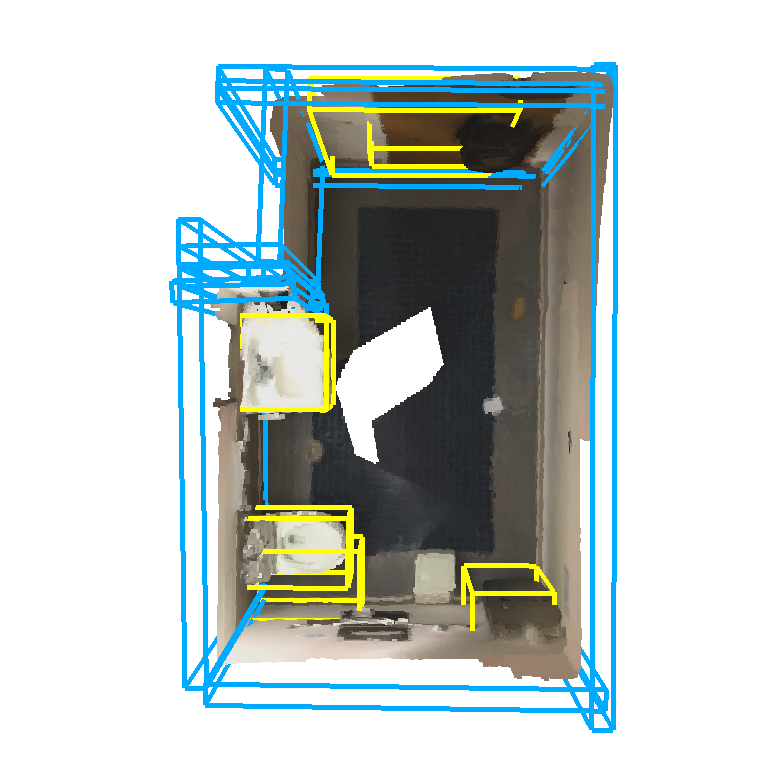} &
     \includegraphics[width=0.12\linewidth, valign=c]{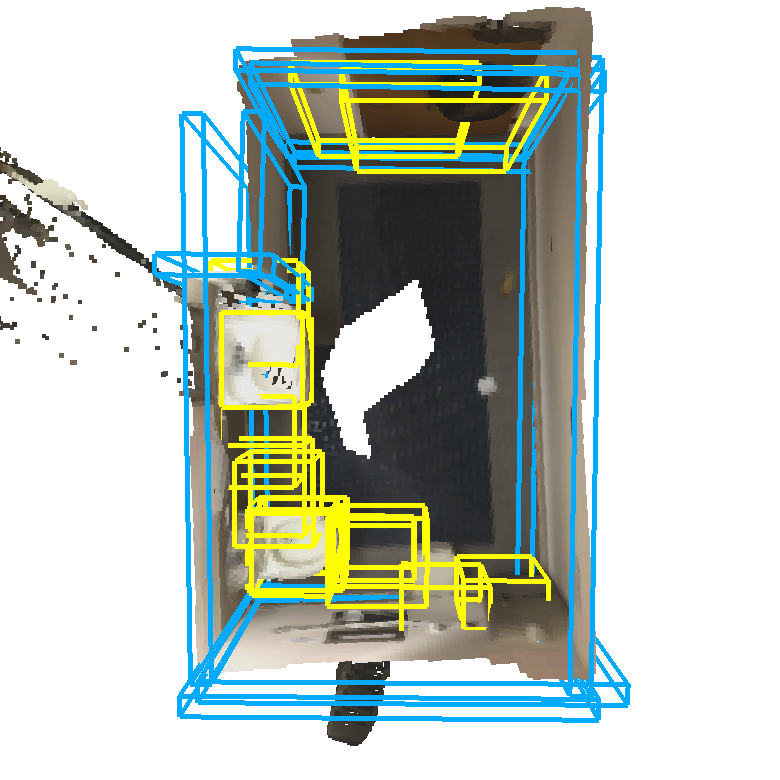} &
     &
     \includegraphics[width=0.12\linewidth, valign=c]{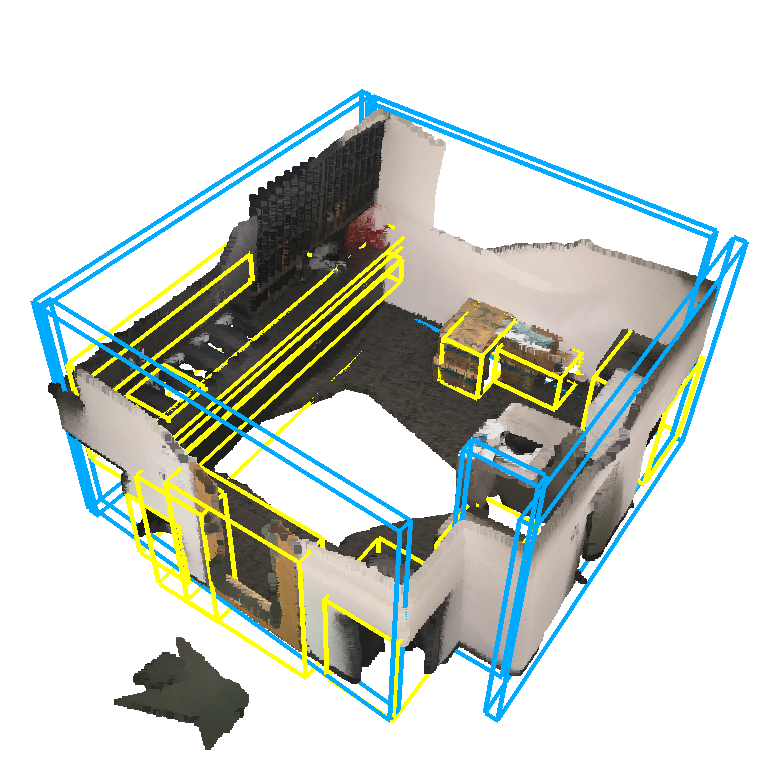} &
     \includegraphics[width=0.12\linewidth, valign=c]{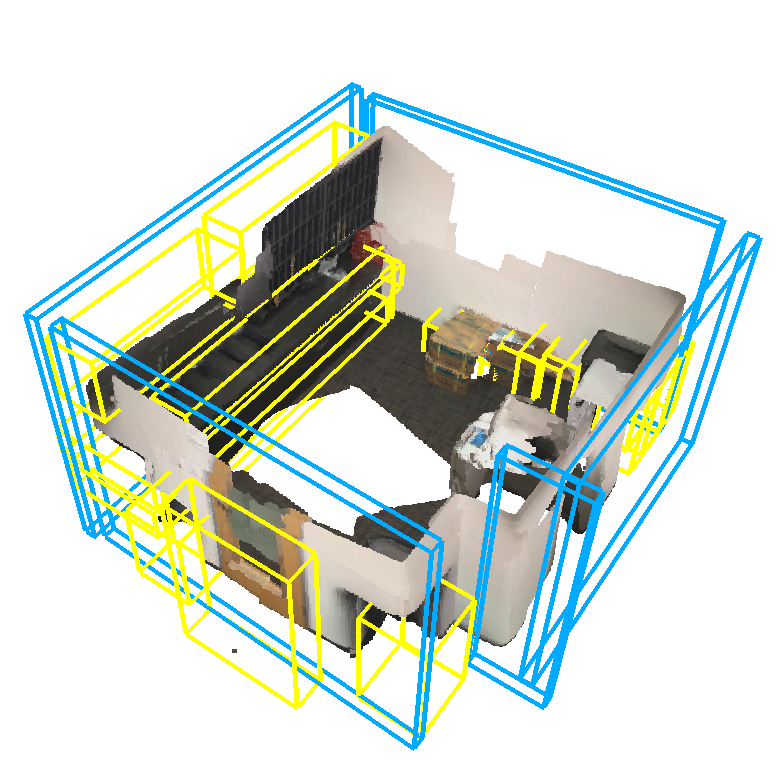} &
     \includegraphics[width=0.12\linewidth, valign=c]{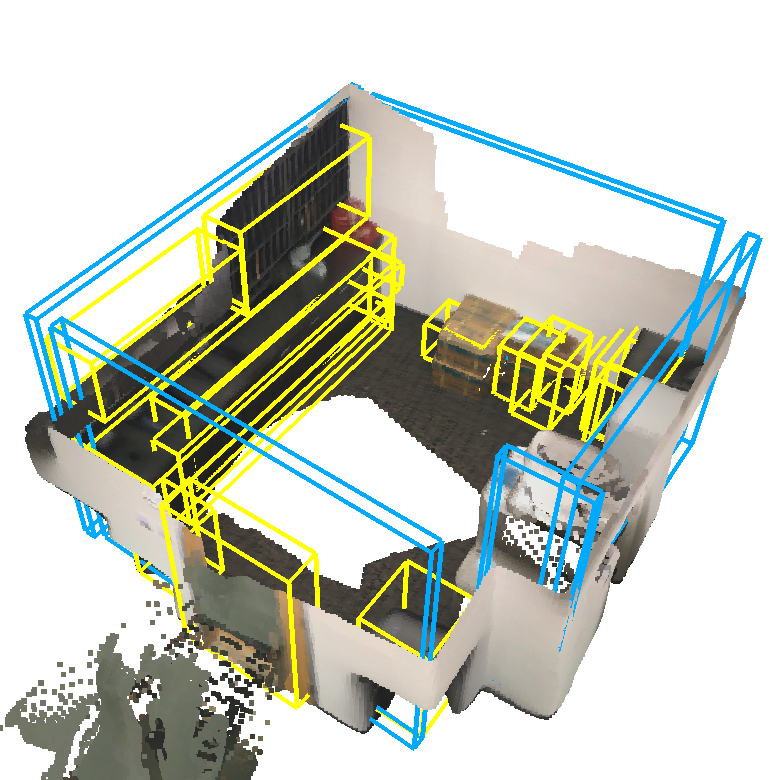} \\
     \rotatebox[origin=c]{90}{\ours{}} & 
     \includegraphics[width=0.12\linewidth, valign=c]{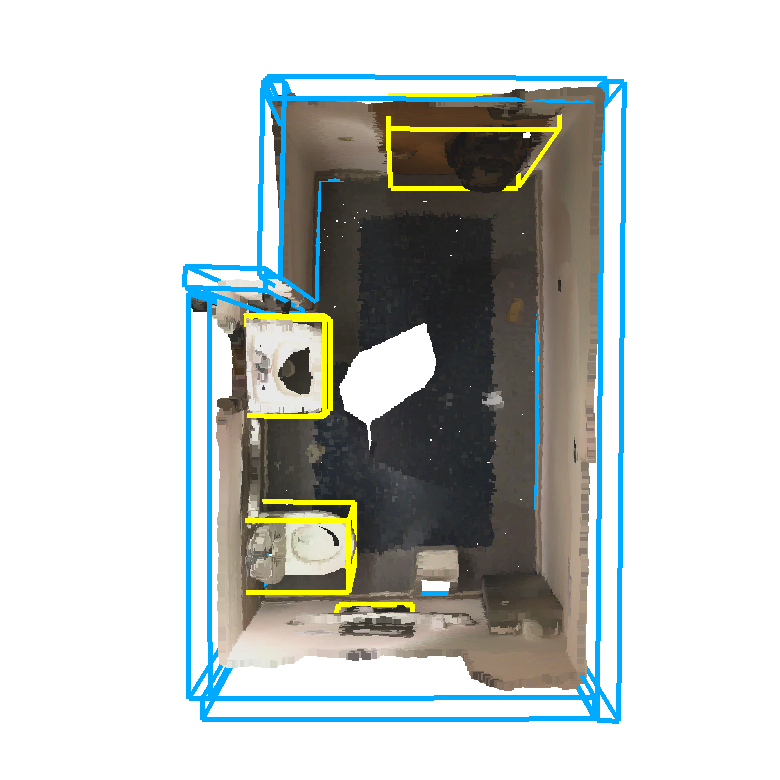} &
     \includegraphics[width=0.12\linewidth, valign=c]{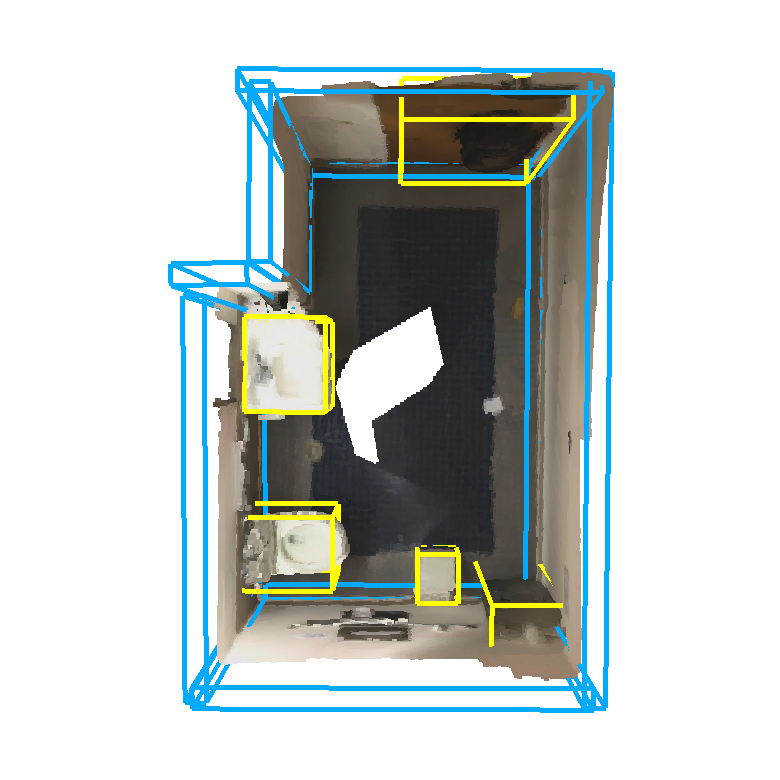} &
     \includegraphics[width=0.12\linewidth, valign=c]{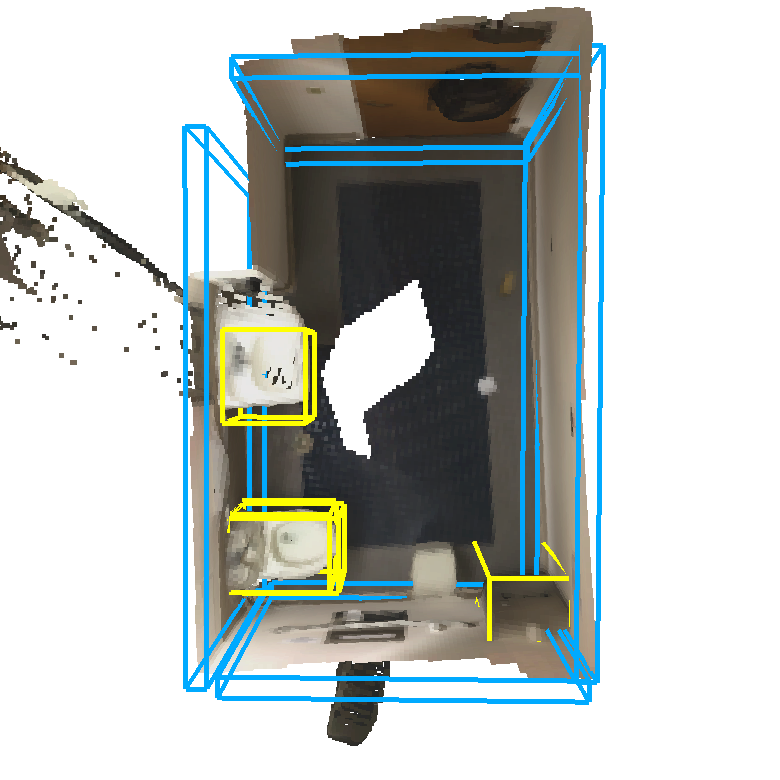} &
     &
     \includegraphics[width=0.12\linewidth, valign=c]{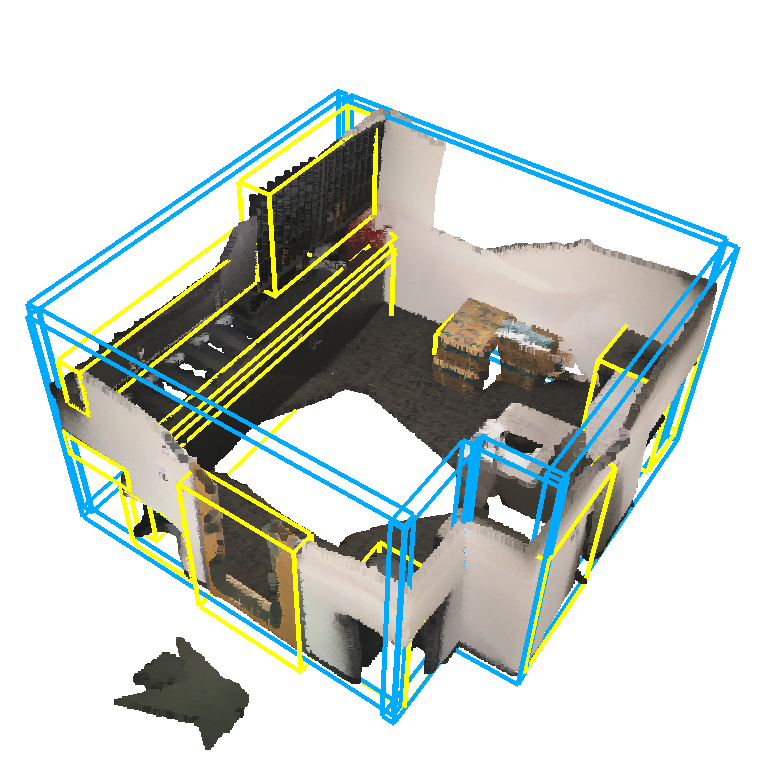} &
     \includegraphics[width=0.12\linewidth, valign=c]{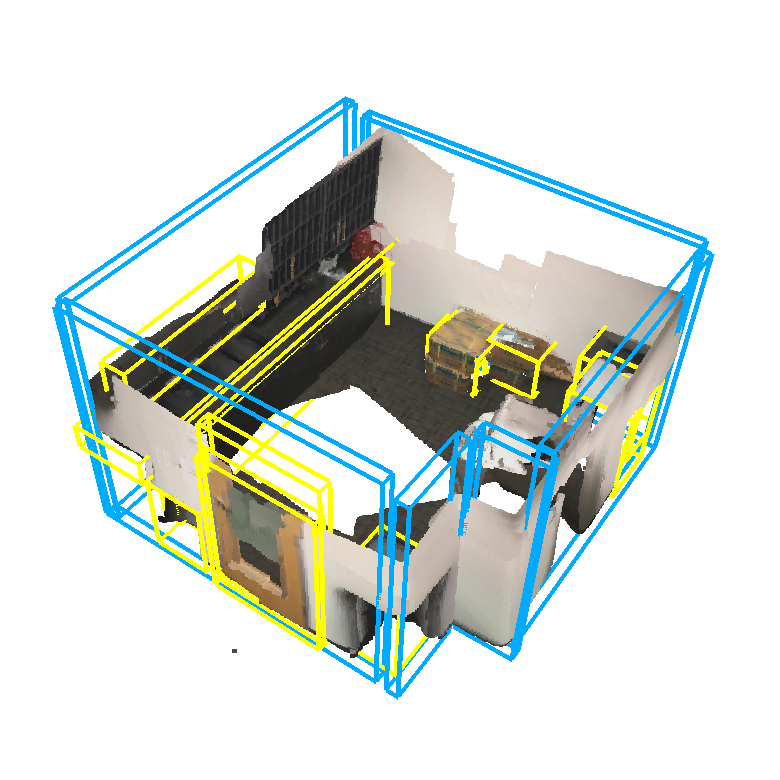} &
     \includegraphics[width=0.12\linewidth, valign=c]{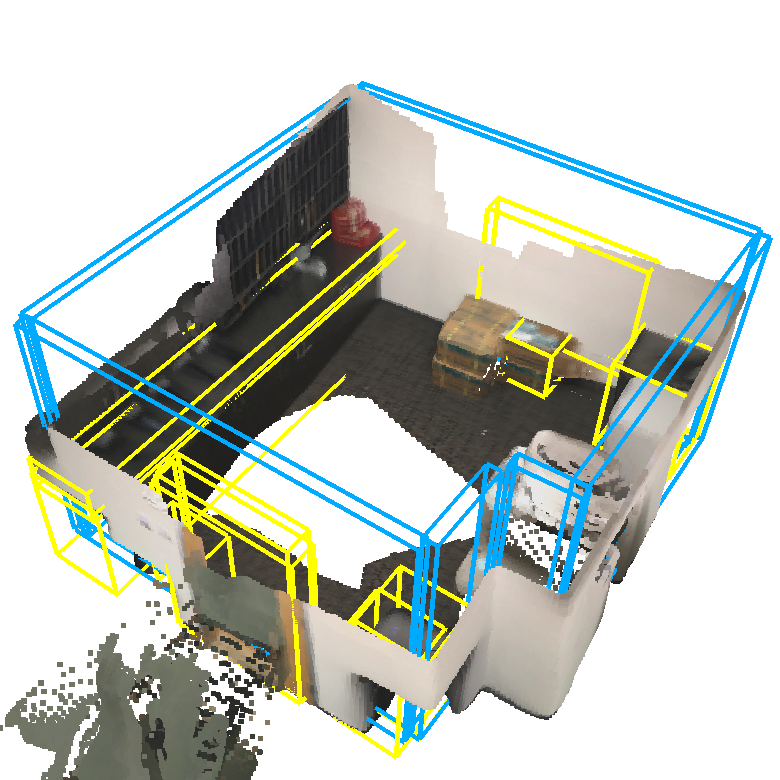} \\
    \rotatebox[origin=c]{90}{\shortstack{GT \\ annotations}} & 
     \includegraphics[width=0.12\linewidth, valign=c]{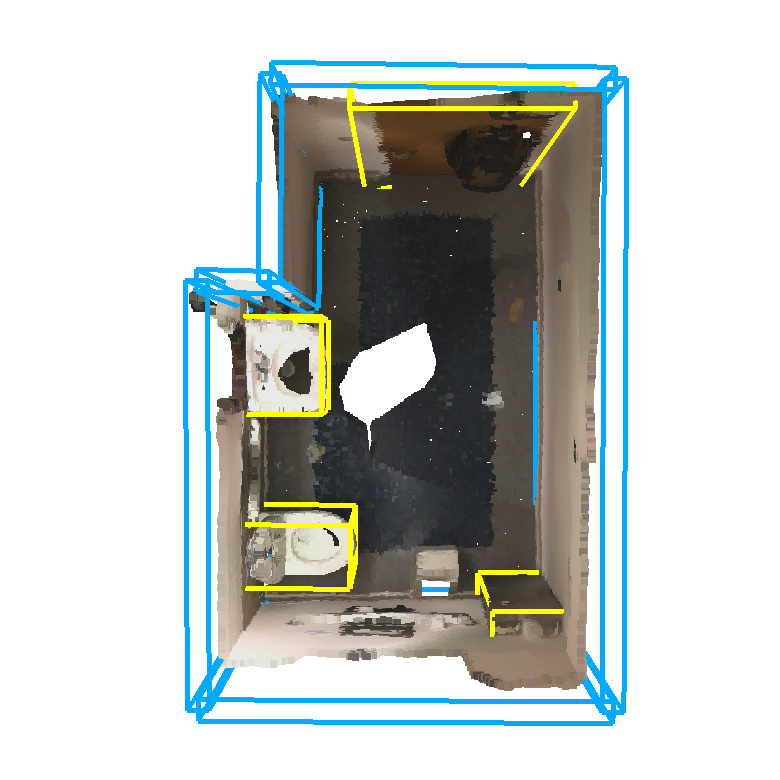} &
     \includegraphics[width=0.12\linewidth, valign=c]{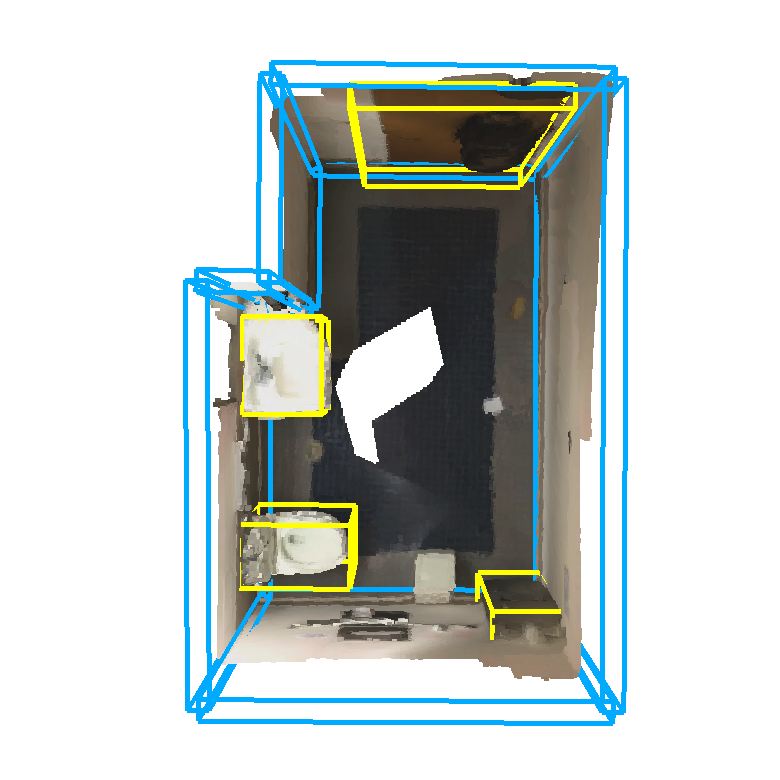} &
     \includegraphics[width=0.12\linewidth, valign=c]{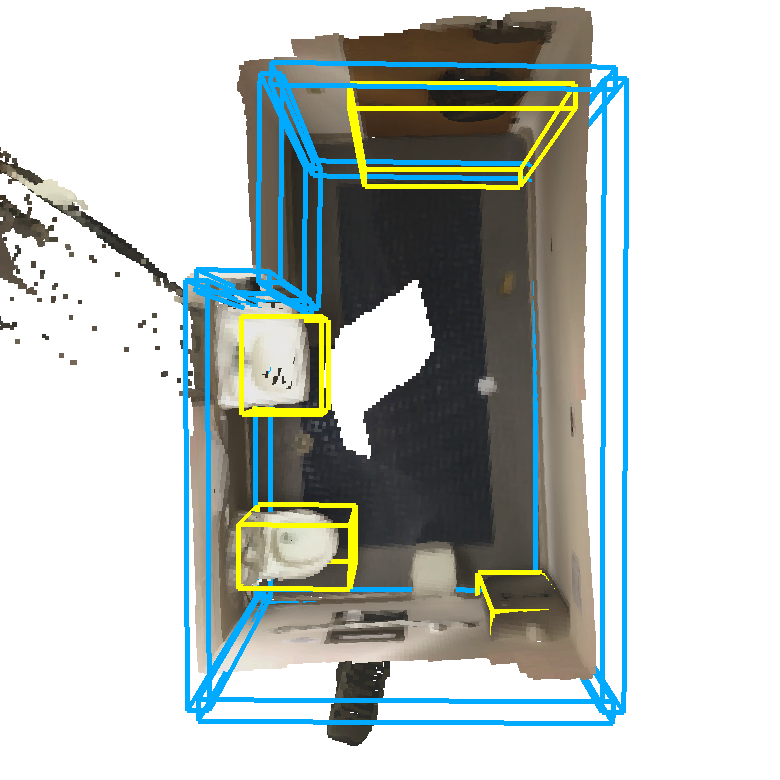} &
     &
     \includegraphics[width=0.12\linewidth, valign=c]{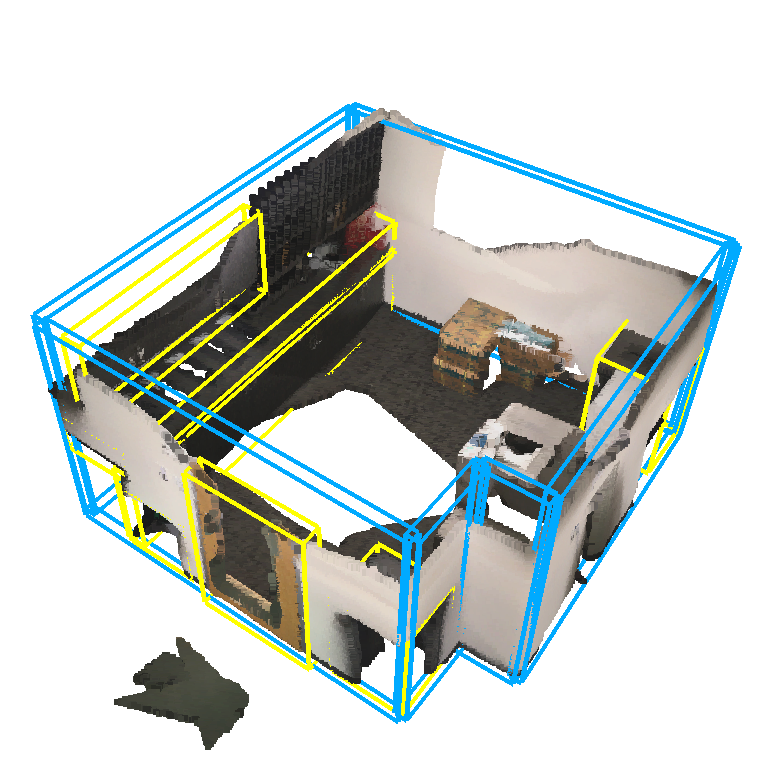} &
     \includegraphics[width=0.12\linewidth, valign=c]{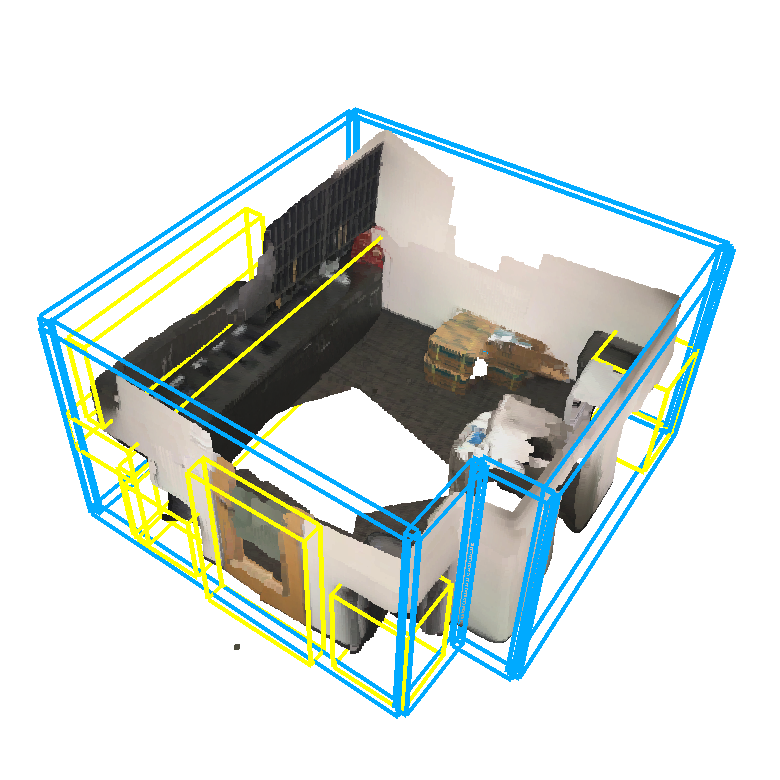} &
     \includegraphics[width=0.12\linewidth, valign=c]{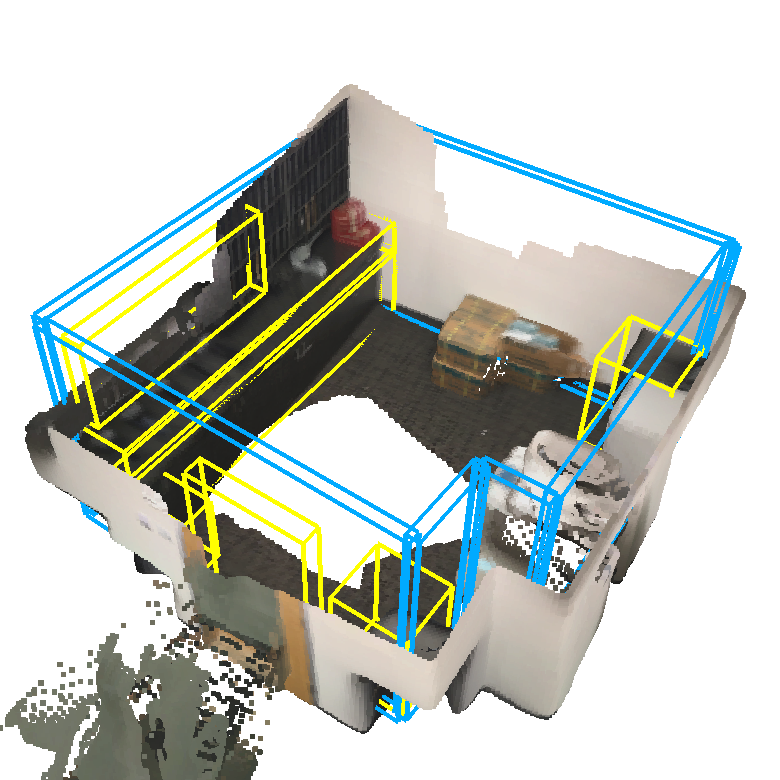} \\
    & \multicolumn{3}{c}{Scene 0146\_02} & & \multicolumn{3}{c}{Scene 0338\_00} \\
    \end{tabular}
    \caption{Ground truth and predicted \textcolor{cyan}{\textbf{layouts}} and \textcolor{yellow}{\textbf{objects}} on ScanNet dataset.}
    \label{fig:qualitative_scannet}
\end{figure*}

\begin{figure*}[t!]
    \centering \small
    \begin{tabular}{cccccccc}
     & GT point clouds & Posed RGB & RGB-only & & GT point clouds & Posed RGB & RGB-only \\
     \rotatebox[origin=c]{90}{PQ~\cite{chen2022pq}} & 
     \includegraphics[width=0.12\linewidth, valign=c]{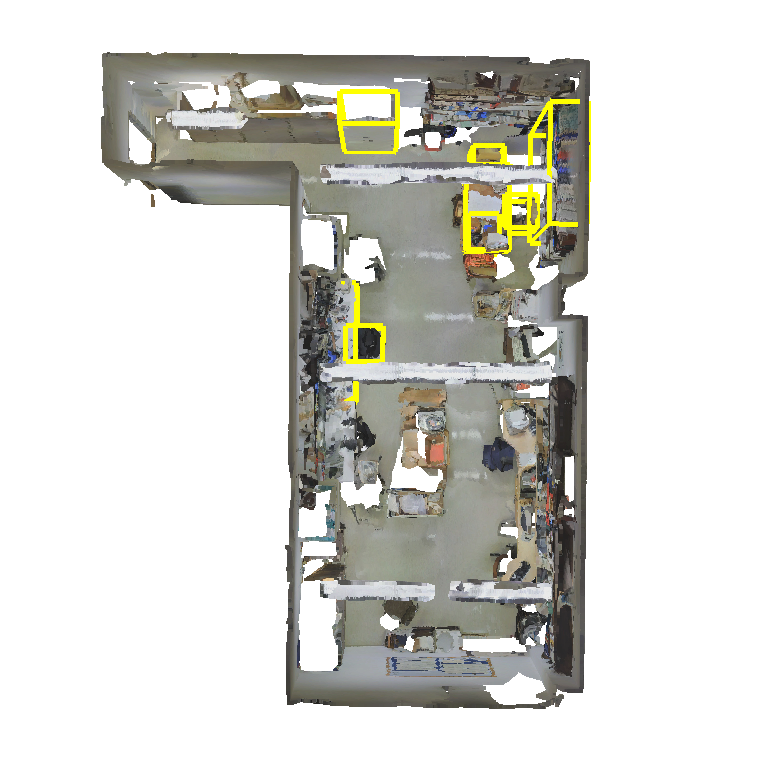} &
     \includegraphics[width=0.12\linewidth, valign=c]{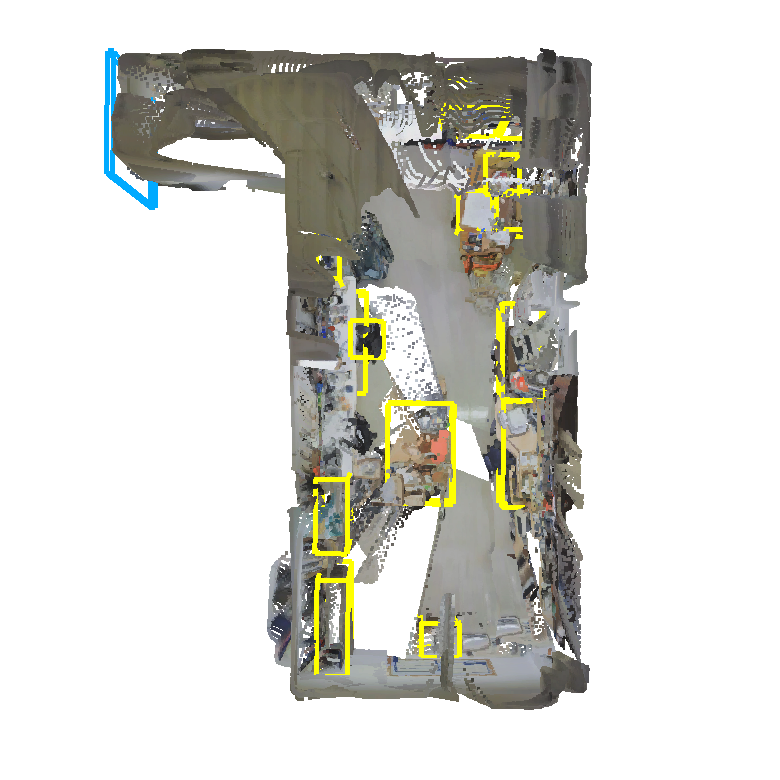} &
     \includegraphics[width=0.12\linewidth, valign=c]{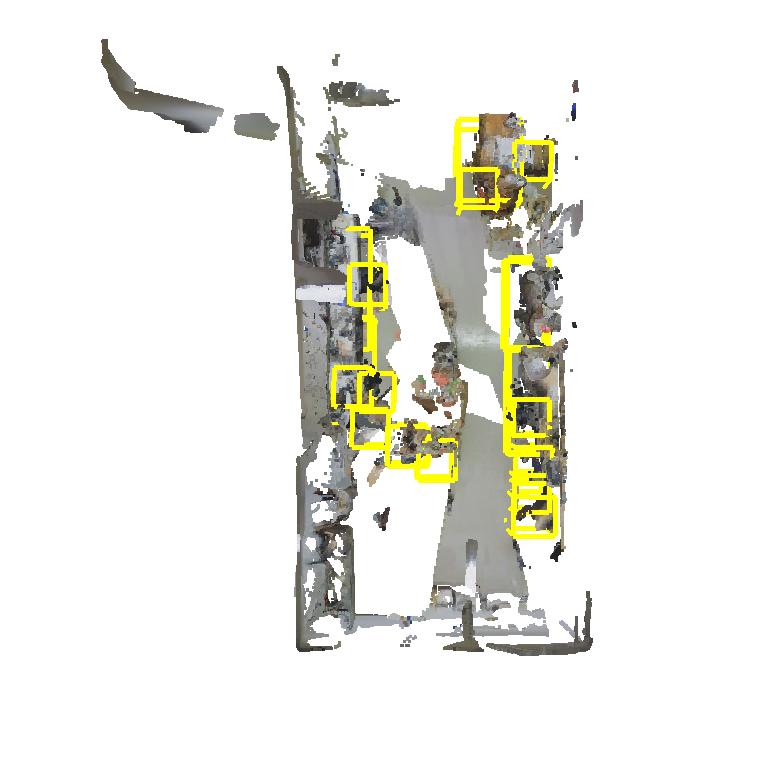} &
     &
     \includegraphics[width=0.12\linewidth, valign=c]{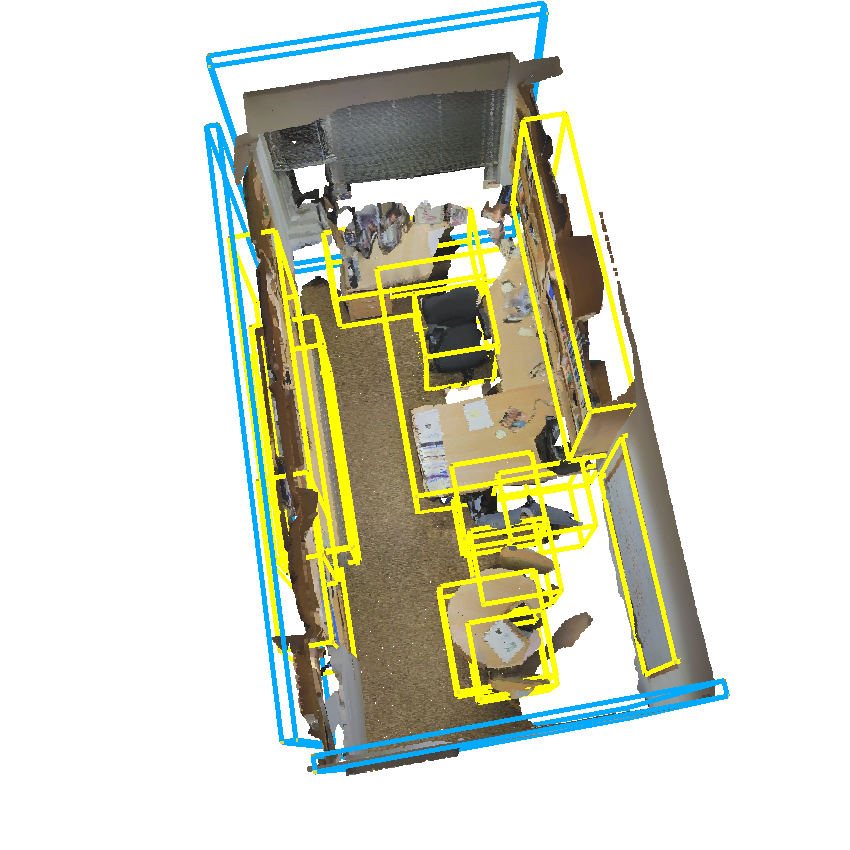} &
     \includegraphics[width=0.12\linewidth, valign=c]{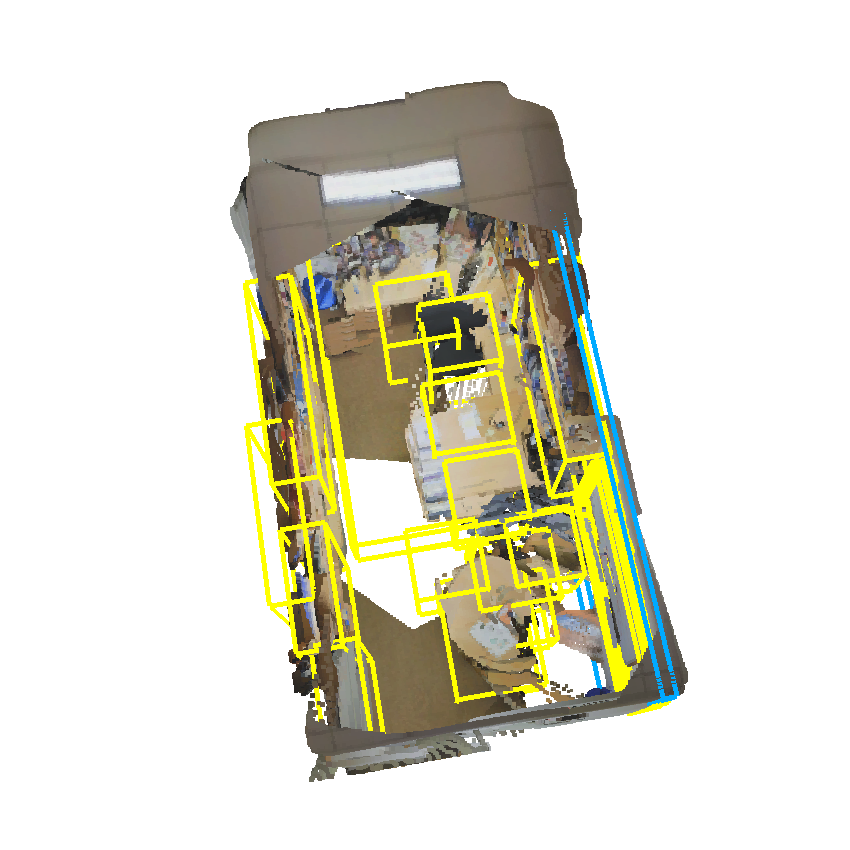} &
     \includegraphics[width=0.12\linewidth, valign=c]{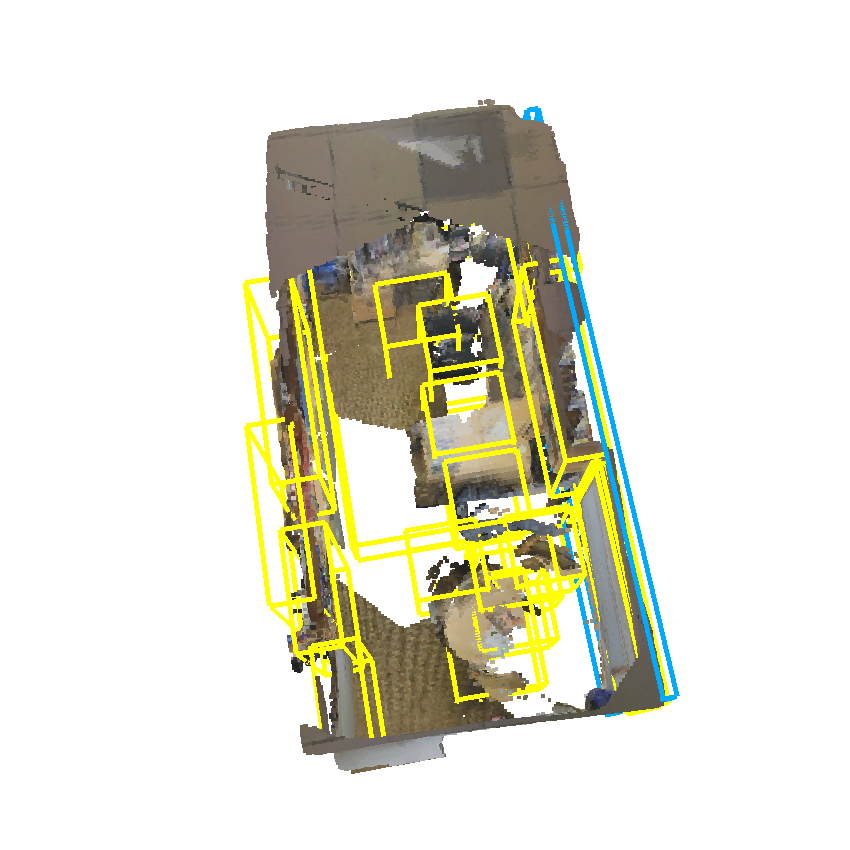} \\
     \rotatebox[origin=c]{90}{\ours{}} &
     \includegraphics[width=0.12\linewidth, valign=c]{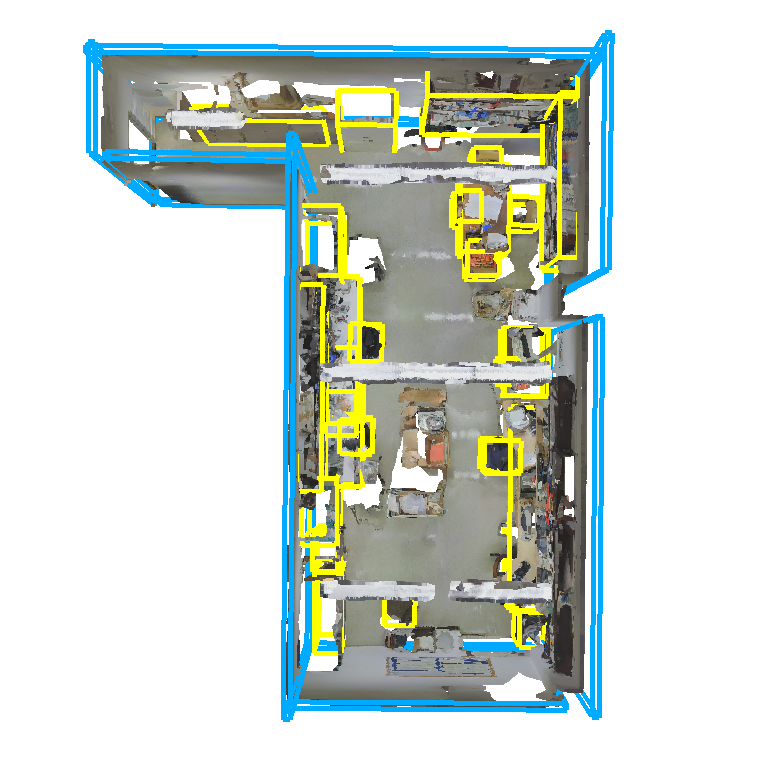} &
     \includegraphics[width=0.12\linewidth, valign=c]{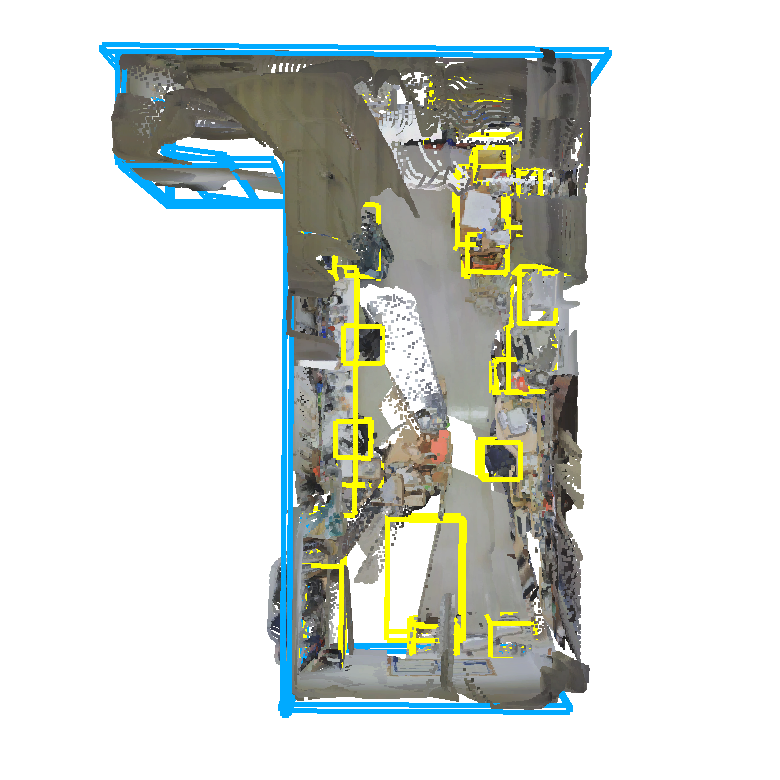} &
     \includegraphics[width=0.12\linewidth, valign=c]{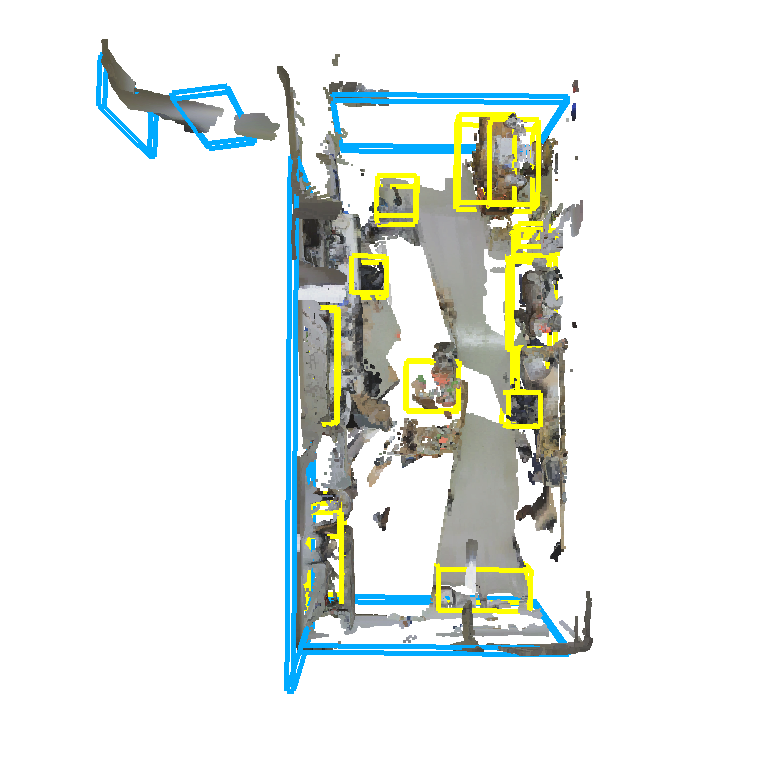} &
     &
     \includegraphics[width=0.12\linewidth, valign=c]{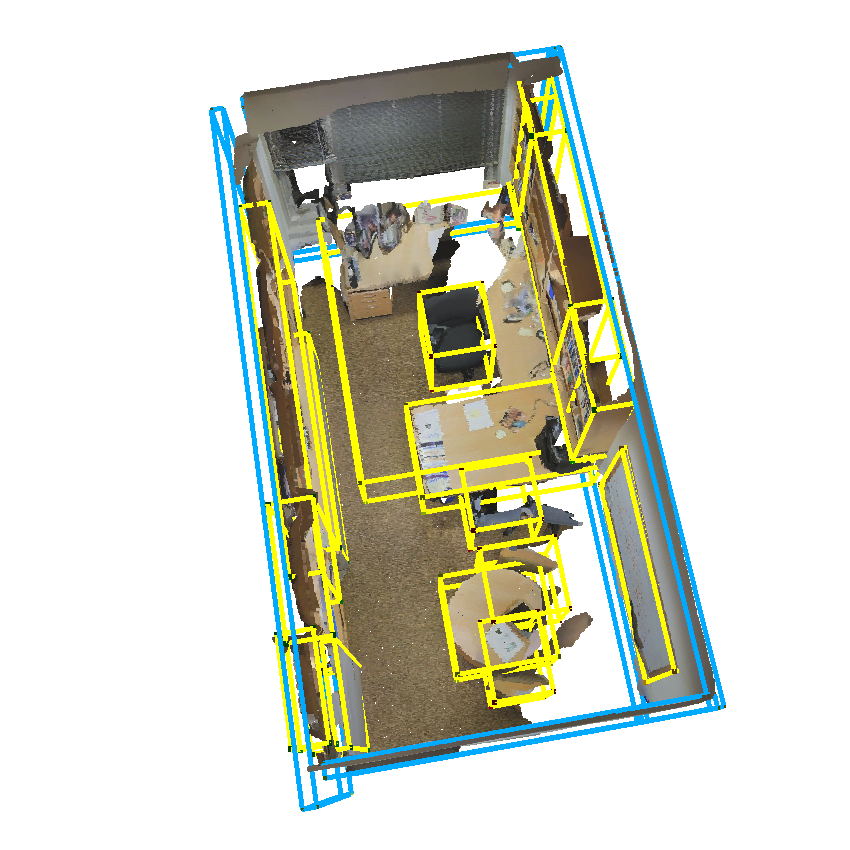} &
     \includegraphics[width=0.12\linewidth, valign=c]{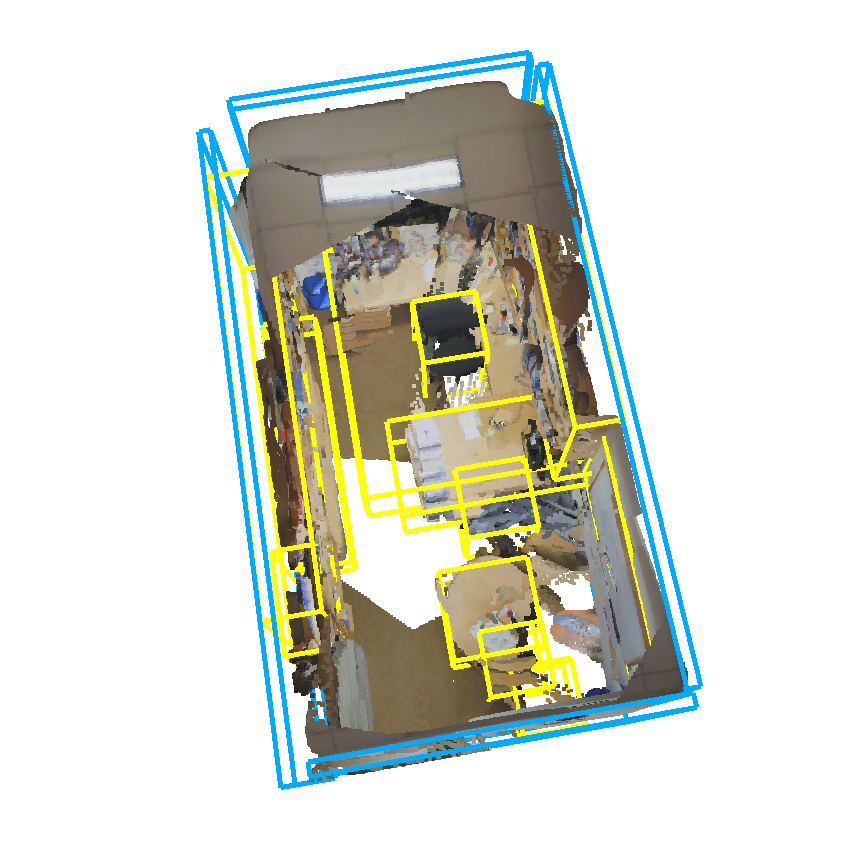} &
     \includegraphics[width=0.12\linewidth, valign=c]{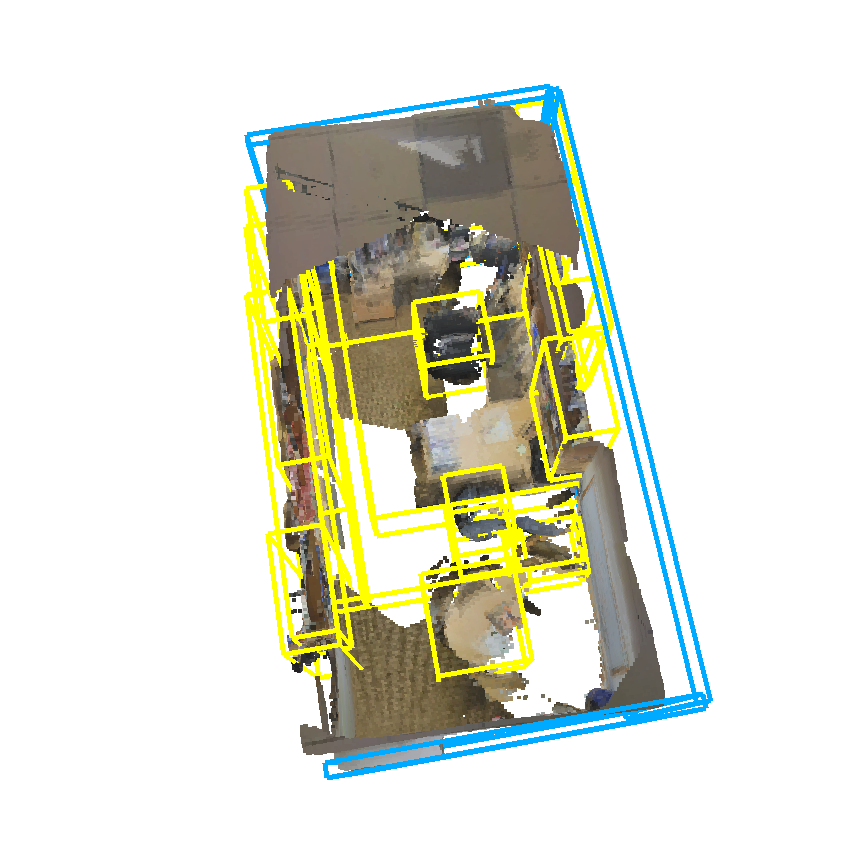} \\
    \rotatebox[origin=c]{90}{\shortstack{GT\\annotations}} & 
     \includegraphics[width=0.12\linewidth, valign=c]{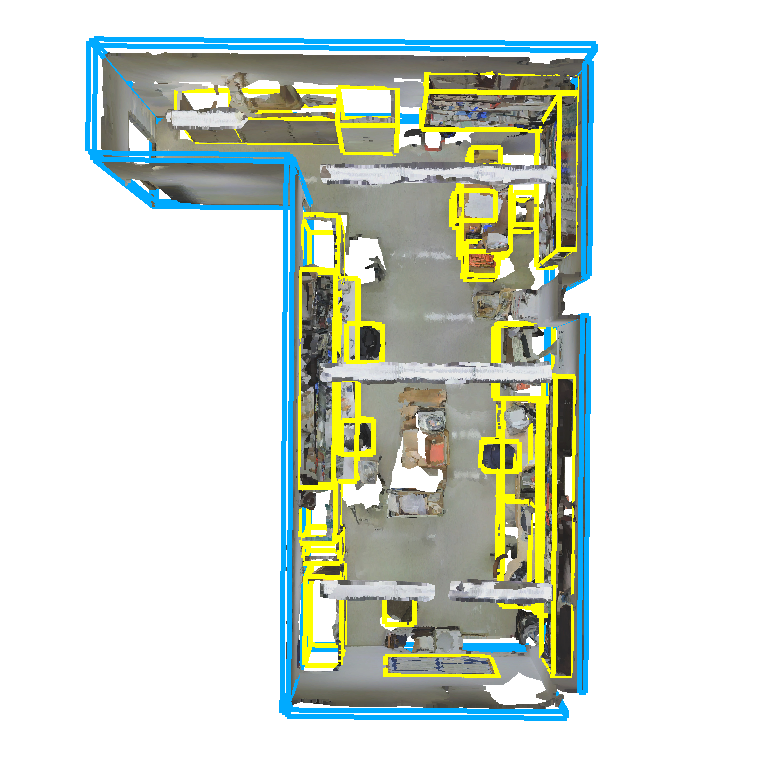} &
     \includegraphics[width=0.12\linewidth, valign=c]{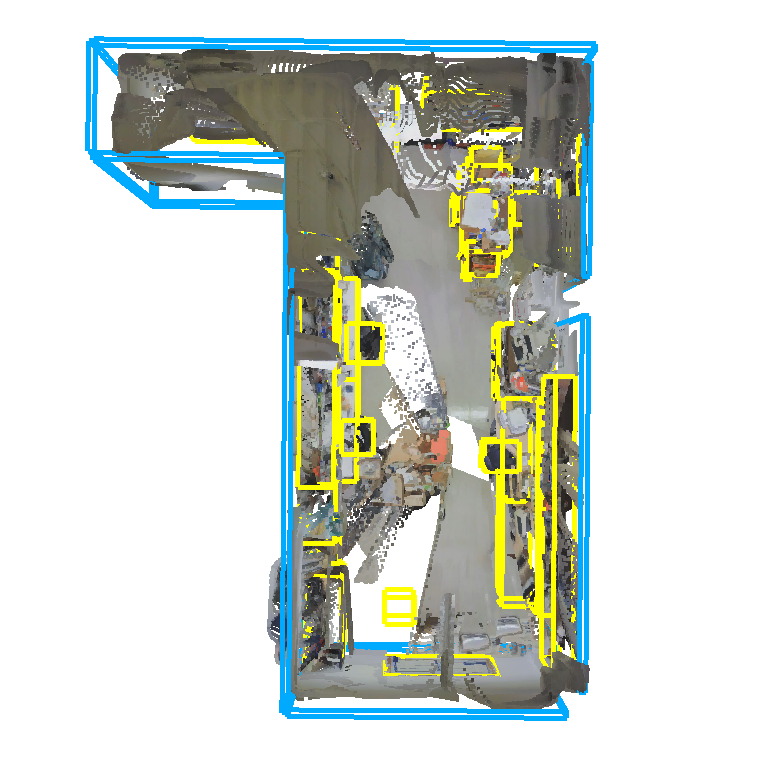} &
     \includegraphics[width=0.12\linewidth, valign=c]{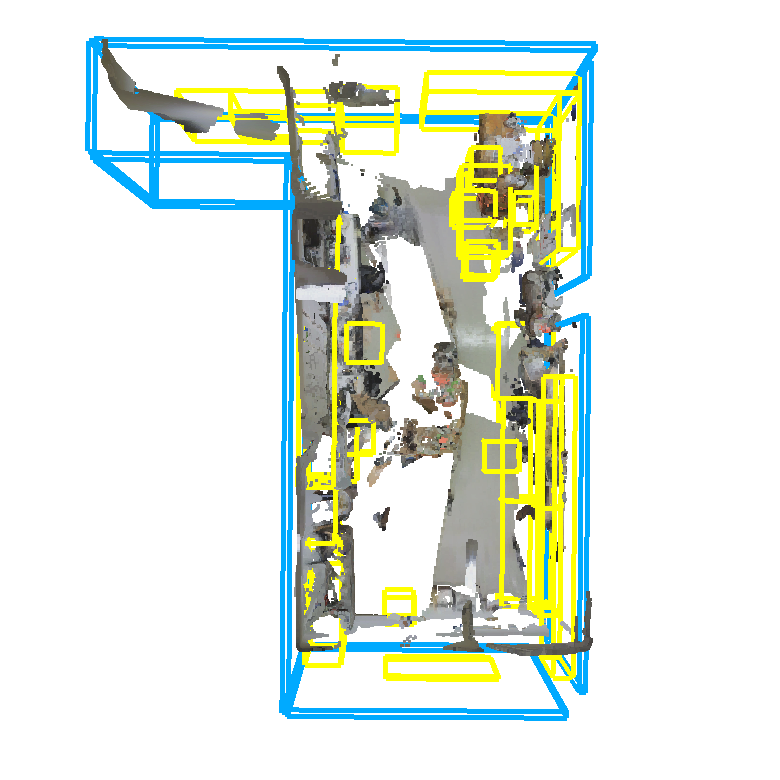} &
     &
     \includegraphics[width=0.12\linewidth, valign=c]{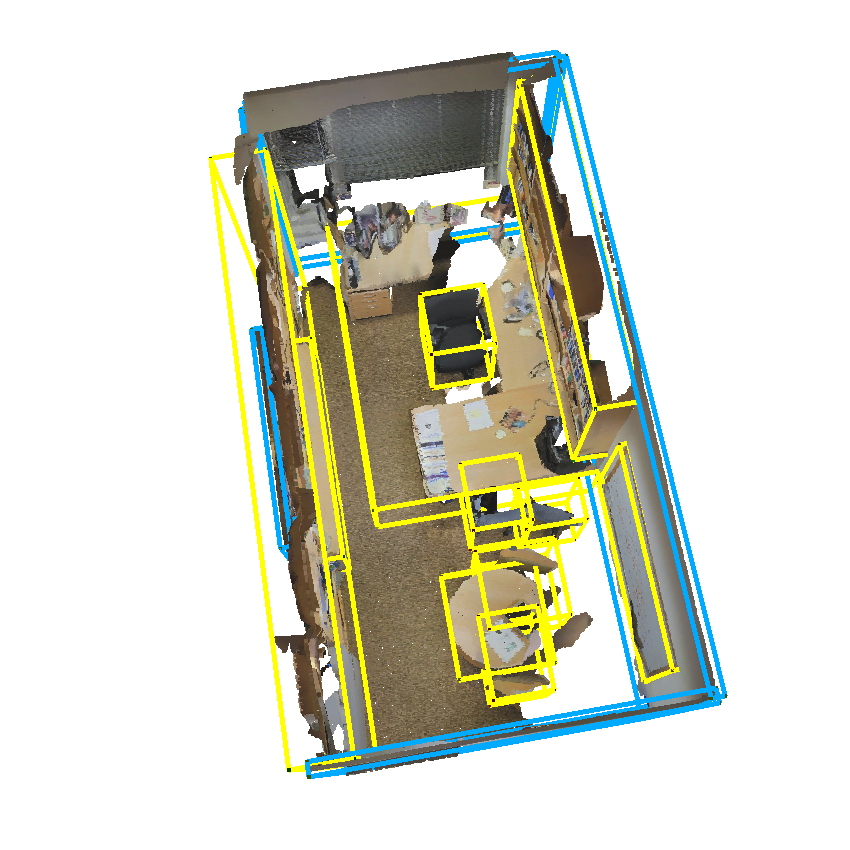} &
     \includegraphics[width=0.12\linewidth, valign=c]{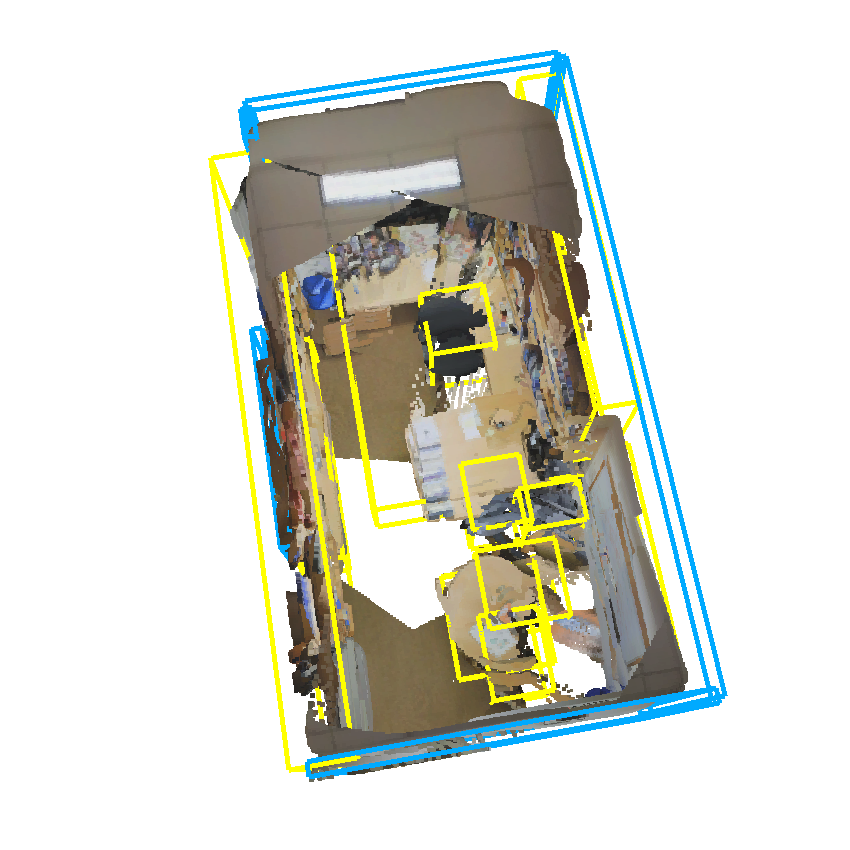} &
     \includegraphics[width=0.12\linewidth, valign=c]{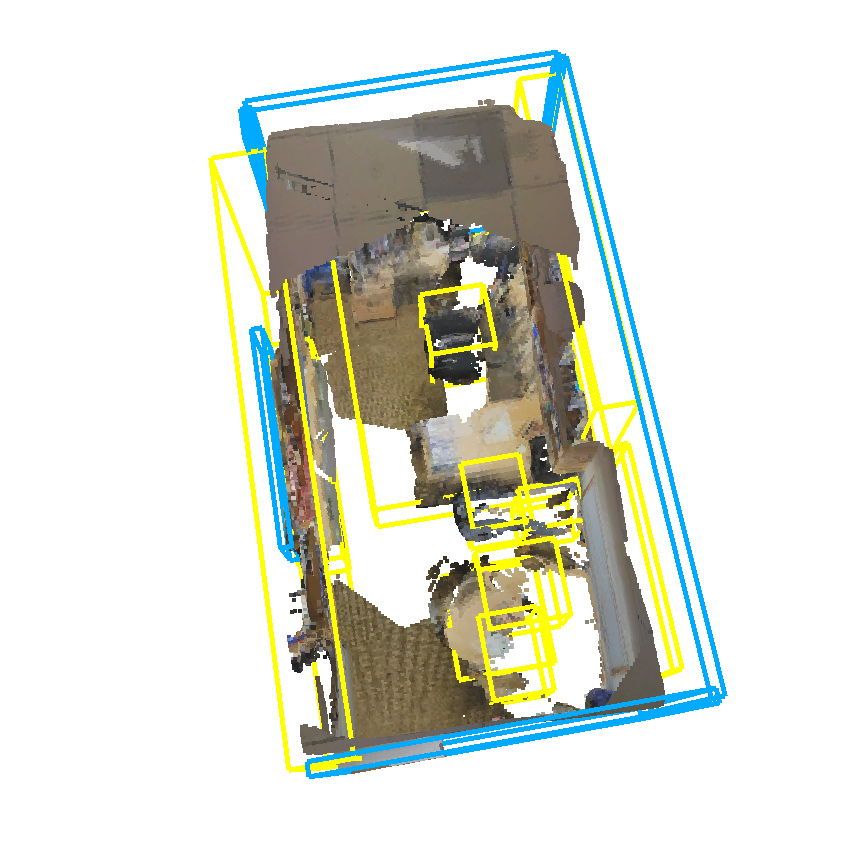} \\
    & \multicolumn{3}{c}{Office 40} & & \multicolumn{3}{c}{Office 6} \\
    \end{tabular}
    \caption{Ground truth and predicted \textcolor{cyan}{\textbf{layouts}} and \textcolor{yellow}{\textbf{objects}} on S3DIS dataset.}
    \label{fig:qualitative_s3dis}
\end{figure*}

\section{RESULTS}

\subsection{Comparison with Prior Approaches}

\inline{Point clouds}. We benchmark the proposed approach in the most common scenario with ground truth point clouds as inputs and report scores on ScanNet and S3DIS (Tab.~\ref{tab:results}), ARKitScenes (Tab.~\ref{tab:results_arkit}) and Structured3D (Tab.~\ref{tab:results_structured3d}). Our method demonstrates superiority in all four benchmarks, with significant performance gains w.r.t. prior state-of-the-art layout estimation methods ($+23.6$ F1 over PQ-Transformer on S3DIS, $+5.8$ and $+4.4$ F1 Omni-PQ on ScanNet and ARKitScenes, $+4.0$ F1@0.25 over SpatialLM on Structured3D). On S3DIS, we report state-of-the-art results among all existing 3D object detection methods. Overall, with consistent improvement over the baseline PQ-Transformer, \ours{} sets a new state-of-the-art in 3D scene understanding.

\begin{table}
\centering
\caption{Results of layout estimation from ground truth points clouds on ARKitScenes.}
\label{tab:results_arkit}
\begin{tabular}{lc}
\toprule
Method & Layout F1 \\
\midrule
PQ~\cite{chen2022pq} & 10.7  \\
Omni-PQ~\cite{gao2023omnipq} & 25.9  \\
\textbf{\ours{}} & \textbf{30.3}  \\
\bottomrule
\end{tabular}
\end{table}

\begin{table}
\centering
\caption{Results of layout estimation and 3D object detection from ground truth point clouds on Structured3D.}
\label{tab:results_structured3d}
\begin{tabular}{lcccc}
\toprule
\multirow{2}{*}{Method} & \multicolumn{2}{c}{Layout} & \multicolumn{2}{c}{Detection} \\
& F1@0.25 & F1@0.5 & F1@0.25 & F1@0.5 \\
\midrule
RoomFormer~\cite{yue2023roomformer} & 70.4 & 67.2 & - & - \\
SceneScript~\cite{avetisyan2024scenescript} & 83.1 & 80.8 & 49.1 & 36.8 \\
SpatialLM~\cite{mao2025spatiallm} & 86.5 & 84.6 & 65.6 & 52.6 \\
\textbf{\ours{}} & \textbf{90.5} & \textbf{89.6} & \textbf{73.9} & \textbf{65.4} \\
\bottomrule
\end{tabular}
\end{table}

\inline{Posed images.} While numerous approaches report 3D object detection performance on ScanNet, we are the first to address layout estimation from posed videos in this popular indoor benchmark. So, to establish a baseline, we combine DUSt3R with PQ-Transformer (denoted as DUSt3R $\rightarrow$ PQ in Tab.~\ref{tab:results}). \ours{} outperforms the baseline in both object detection and layout estimation $+11.1$ F1 on ScanNet, $+27.7$ F1 on S3DIS. For apple-to-apple comparison, we separate existing approaches based on the use of depth data for training. \ours{} clearly dominates in the depth-agnostic category and even outperforms recent depth-aware methods in terms of mAP@0.5.

\inline{Unposed images.} In the third track, there are no predecessors reporting scene understanding results on real scans. Respectively, we also obtain reference numbers with a combination of DUSt3R $\rightarrow$ PQ. As could be expected, DUSt3R $\rightarrow$ \ours{} outperforms the baseline by a large margin. More surprisingly, predicted layouts appear to be more accurate than the ones produced by DUSt3R $\rightarrow$ PQ \textit{with} poses, which actually means that we can drop one input modality (camera poses) yet achieve the same quality of layout estimation.

\subsection{Ablation Experiments}

\inline{Inference time}. In Tab.~\ref{tab:time}, we compare the efficiency of \ours{} and other methods that report both layout estimation and object detection on ScanNet and S3DIS. As can be seen, our approach is orders of magnitude faster than LLM-based SpatialLM and 4x faster than PQ-Transformer. 

\begin{table}
\centering
\caption{Inference time (ms) with posed images from ScanNet and S3DIS.}
\label{tab:time}
\begin{tabular}{lcc}
\toprule
Method & ScanNet & S3DIS \\
\midrule
PQ~\cite{chen2022pq} & 217 & 256 \\
SpatialLM~\cite{mao2025spatiallm} & 7935 & 7976 \\
\textbf{\ours{}} & \textbf{49} & \textbf{79} \\
\bottomrule
\end{tabular}
\end{table}

Our main architectural choices are driven by efficiency, including adapting a TR3D-like backbone. To prove them, we experiment with the arguably more accurate recent 3D object detection model, UniDet3D~\cite{kolodiazhnyi2025unidet3d}, that produces a set of 3D predictions with a transformer decoder. Following the internal logic of the UniDet3D's processing pipeline, we add a layout head that predicts walls as 4$\times$3D offsets. In Tab.~\ref{tab:ablation_unidet3d}, we compare UniDet3D against \ours{} with the same wall parameterization and the final version of \ours{} with the proposed parameterization. Evidently, \ours{} is 1.7x faster while improving the layout F1 score by $+4.4$. 3D object detection results are inferior to the ones of UniDet3D on ScanNet and superior on S3DIS ($+1.2$ mAP@0.5 reported in Tab.~\ref{tab:results}), so an ultimate leader cannot be identified. Overall, we claim that with our architecture, we reach a decent balance of efficiency and accuracy.

\inline{Pose estimation.} For videos, e.g., from ScanNet, an indoor SLAM seems to be an obvious candidate to estimate poses and create a map of the scanned environment. In Tab.~\ref{tab:ablation_slam}, we report scores obtained with DUSt3R~\cite{wang2024dust3r} and a solid SLAM baseline, DROID-SLAM~\cite{teed2021droidslam}. Evidently, DUSt3R produces far more precise geometry, leading to a 2x improvement in both layout estimation and 3D object detection accuracy.

\begin{table}
\centering
\caption{Results of layout estimation and object detection from unposed images on ScanNet with different pose estimation methods.}
\label{tab:ablation_slam}
\begin{tabular}{lccc}
\toprule
\multirow{2}{*}{Method} & Layout & \multicolumn{2}{c}{Detection}  \\
& F1 & mAP@0.25 & mAP@0.5 \\
\midrule
DROID-SLAM $\rightarrow$ TUN3D & 23.2 & 23.9 & 9.5 \\
DUSt3R $\rightarrow$ TUN3D & \textbf{46.5} & \textbf{44.0} & \textbf{20.7} \\
\bottomrule
\end{tabular}
\end{table}

\begin{table}
\centering
\caption{Results of layout estimation and object detection from posed images on ScanNet with varying number of images.}
\label{tab:ablation_num_frames}
\begin{tabular}{lccc}
\toprule
\multirow{2}{*}{\# Images} & Layout & \multicolumn{2}{c}{Detection}  \\
& F1 & mAP@0.25 & mAP@0.5 \\
\midrule
15                   & 43.3 & 47.1 & 26.5 \\
25                   & 50.1 & 52.2 & 33.1 \\
35                   & \textbf{55.2} & 56.1 & 35.0 \\
45                   & \textbf{55.2} & \textbf{57.4} & \textbf{35.6} \\
\bottomrule
\end{tabular}
\end{table}

\inline{Number of images.} To identify the sufficient level of coverage, we vary the number of input images. According to the Tab.~\ref{tab:ablation_num_frames}, scores increase with the number of frames. We use 45 frames in the final version of \ours{}, which is on par with our competitors: ImVoxelNet~\cite{rukhovich2022imvoxelnet} uses 50 images, NeRF-Det~\cite{xu2023nerfdet} -- 100 images.

\inline{Number of $z$-quantiles.} In Tab.~\ref{tab:ablation_z}, we identify the best level of detail when encoding spatial information about the scene. Our model demonstrates reasonable performance even without such information. Yet, 10 $z$-quantiles bring $+5.2$ F1 on ScanNet and $+4.3$ F1 on S3DIS, which is a noticeable improvement obtained with a negligible computation overhead having no effect on the overall inference time.

\begin{table}
\centering
\caption{Results of layout estimation from posed images on ScanNet and S3DIS with varying number of $z$-quantiles.}
\label{tab:ablation_z}
\begin{tabular}{lcc}
\toprule
\# $z$-quantiles & ScanNet & S3DIS \\
\midrule
0 & 50.0 & 33.6 \\
1 & 53.3 & 33.3 \\
2 & 53.8 & 35.3 \\
5 & 55.1 & 36.4 \\
10 & \textbf{55.2} & \textbf{37.9} \\
\bottomrule
\end{tabular}
\end{table}

\inline{Wall parameterization.} To prove the efficiency of our proposed wall parameterization, we test it against other options in Tab.~\ref{tab:ablation_layout_params}. Our parameterization is not only expressed with as few as 5 parameters, but also improves layout F1 score by $+1.3$ on ScanNet.

\begin{table}
\centering
\caption{Results of layout estimation from posed images on ScanNet with different wall parameterizations.}
\label{tab:ablation_layout_params}
\begin{tabular}{lcc}
\toprule
Wall parameterization & \# parameters & Layout F1 \\
\midrule
PQ~\cite{chen2022pq} & 8    & 51.2 \\
4$\times$3D offsets & 12 & 53.2  \\
2$\times$3D offsets + height & 7 & 53.9  \\
2$\times$2D offsets + height & 5 & \textbf{55.2}  \\
\bottomrule
\end{tabular}
\end{table}

\begin{table}
\centering
\caption{Results of layout estimation and object detection from ground truth point clouds on ScanNet with different backbone architectures.}
\label{tab:ablation_unidet3d}
\begin{tabular}{lcccc}
\toprule
\multirow{2}{*}{Method} & Wall & Layout & Detection & Time, \\
& param. & F1 & mAP@0.25 & ms\\
\midrule
UniDet3D+layout & 4$\times$3D offsets & 61.8 & \textbf{77.0} & 213 \\
\ours{} & 4$\times$3D offsets & 63.2 & 72.7 & \textbf{127} \\
\textbf{\ours{}} & ours & \textbf{66.6} & 72.7 & \textbf{127} \\
\bottomrule
\end{tabular}
\end{table}

%\begin{table}[h!]
%\centering
%\caption{Results of layout estimation and object detection from unposed images on ScanNet with different training datasets. Model trained on Structured3D is significantly worse than models trained on ScanNet. This confirms the hypothesis that modern synthetic datasets for 3D scene understanding cannot yet be fully utilized for real-world domain applications. Moreover, training on ground truth scans is worse than training on scans from unposed images, which shows the importance of matching domains at the training and validation stage even when we work with the same dataset.}
%\label{tab:ablation_num_frames}
%\begin{tabular}{lccc}
%\toprule
%\multirow{2}{*}{Training dataset} & Layout & \multicolumn{2}{c}{Detection}  \\
%& F1 & mAP@0.25 & mAP@0.5 \\
%\midrule
%Structured3D                   & 7.6 & - & - \\
%ScanNet (gt point clouds)      & 41.7 & 34.9 & 14.0 \\
%ScanNet (unposed images)       & \textbf{46.5} & \textbf{44.0} & \textbf{20.7} \\
%\bottomrule
%\end{tabular}
%\end{table}

\section{CONCLUSION}

We introduced \ours{}, the first method of joint layout prediction and 3D object detection in real scans that takes multi-view images as an input. Besides, \ours{} is trained without ground-truth camera poses or depths, hence relaxing the input data requirements to casually captured images or videos. To achieve this, we developed a lightweight sparse-convolutional model with two single-task heads and proposed a novel and effective layout parameterization. Through experiments across multiple benchmarks and various data modalities, we proved that \ours{} sets a new state-of-the-art in joint layout estimation and 3D object detection from ground-truth point clouds, posed and unposed images, marking a new milestone in holistic indoor scene understanding.


%%%%%%%%%%%%%%%%%%%%%%%%%%%%%%%%%%%%%%%%%%%%%%%%%%%%%%%%%%%%%%%%%%%%%%%%%%%%%%%%

\bibliographystyle{IEEEtran}
\bibliography{icra}

\end{document}